\documentclass[journal,onecolumn]{IEEEtran}
\usepackage{color,graphicx,epsfig,epsf,latexsym,theorem,hhline,amsmath,amssymb,array}

\usepackage[ruled]{algorithm2e}

\makeatletter
\def\ps@headings{%
\def\@oddhead{\mbox{}\scriptsize\rightmark \hfil \thepage}%
\def\@evenhead{\scriptsize\thepage \hfil \leftmark\mbox{}}%
\def\@oddfoot{}%
\def\@evenfoot{}}
\makeatother
\newcommand {\mymarginpar}[1]{\marginpar{#1}}
\renewcommand {\marginpar}[1]{}

\def\_{\rule{.3em}{.15ex}}      

\newcommand {\bearn}{\begin{eqnarray*}}
\newcommand {\eearn}{\end{eqnarray*}}

\newtheorem{definition}{Definition}
\newtheorem{property}[definition]{Property}
\newtheorem{proposition}[definition]{Proposition}
\newtheorem{lemma}[definition]{Lemma}
\newtheorem{theorem}[definition]{Theorem}
\newtheorem{corollary}[definition]{Corollary}
\newtheorem{example}[definition]{Example}
\newtheorem{remark}[definition]{Remark}

\newtheorem{fact}[definition]{Fact}

\newcommand {\bfig}[2] {\begin{figure}
  \centering
  \includegraphics[width=#2]{#1}}
\newcommand {\brotatefig}[2] {\begin{figure}[htbp]
                        \centerline {
                         \epsfig{figure={#1},clip=,angle=-90,width={#2}}}}

\newcommand {\bfigfirst}[2] {\begin{figure}[h]
                        \centerline {
                        \setlength{\epsfxsize}{#2}
                        \epsffile{#1}}}
\newcommand {\efig}[2]{ \caption{#2}
                        \label{fig:#1}
                        \end{figure}
                        \mymarginpar{fig:#1}}
\newcommand {\erotatefig}[2]{ \caption{#2}
                        \label{fig:#1}
                        \end{figure}
                        \mymarginpar{fig:#1}}
\newcommand {\rfig}[1]{Figure \ref{fig:#1}}

\newcommand {\btab}[1]{
                       \begin{table}
                       \centering
                       \begin{tabular}{#1}}
\newcommand {\etab}[3] {
                       \end{tabular}
                       \caption[#3]{#2}
                       \label{tab:#1}
                       \end{table}
                       \mymarginpar{tab:#1}
                       \vspace{.1in}}

\newcommand {\btabular}[1]{\begin{center}
                       \begin{tabular}{#1}}
\newcommand {\etabular}{\end{tabular}
                       \end{center}}

\newcommand {\bdefin}[1]{\begin{definition}
                      \mymarginpar{def:#1}
                      \label{def:#1} }
\newcommand {\edefin}       {\end{definition}}
\newcommand {\rdef}[1]{Definition \ref{def:#1}}

\newcommand {\bpro}[1]{\begin{property}
                      \mymarginpar{pro:#1}
                      \label{pro:#1} }
\newcommand {\epro}   {\end{property}}

\newcommand {\bprop}[1]{\begin{proposition}
                      \mymarginpar{prop:#1}
                      \label{prop:#1} }
\newcommand {\eprop}       {\end{proposition}}
\newcommand {\rprop}[1]{Proposition \ref{prop:#1}}

\newcommand {\blem}[1]{\begin{lemma}
                      \mymarginpar{lem:#1}
                      \label{lem:#1} }
\newcommand {\elem}   {\end{lemma}}
\newcommand {\rlem}[1]{Lemma \ref{lem:#1}}

\newcommand {\bthe}[1]{\begin{theorem}
                      \mymarginpar{the:#1}
                      \label{the:#1} }
\newcommand {\ethe}   {\end{theorem}}
\newcommand {\rthe}[1]{Theorem \ref{the:#1}}

\newcommand {\bproof}{\noindent {\bf Proof.} \ }
\newcommand {\eproof} {\hspace*{\fill}~\mbox{\rule[0pt]{1.3ex}{1.3ex}}}

\newcommand {\bcor}[1]{\begin{corollary}
                      \mymarginpar{cor:#1}
                      \label{cor:#1} }
\newcommand {\ecor}   {\end{corollary}}
\newcommand {\rcor}[1]{Corollary \ref{cor:#1}}

\newcommand {\bax}[1]{\begin{axiom}
                      \mymarginpar{ax:#1}
                      \label{ax:#1} }
\newcommand {\eax}       {\vspace{-.1in} \end{axiom}}

\newcommand {\bex}[2]{\vspace{.1in}
                      \begin{example}
                      \mymarginpar{ex:#1}
                       {\bf #2}
                      \label{ex:#1} }
\newcommand {\eex}       {\end{example} \vspace{.3cm} }

\newcommand {\brem}[1]{\begin{remark}
                      \mymarginpar{rem:#1}
                      \label{rem:#1} \em }
\newcommand {\erem}   {\end{remark}}

\newcommand {\beq}[1]{\mymarginpar{eq:#1}
                      \begin{equation}
                      \label{eq:#1} }

\newcommand {\beqno}[1]{\mymarginpar{eq:#1}
                      \begin{eqnarray}
                      \nonumber}

\newcommand {\eeq}       {\end{equation}}
\newcommand {\eeqno}       { && \end{eqnarray}}
\newcommand {\req}[1]{(\ref{eq:#1})}

\newcommand {\bear}[1]{\mymarginpar{eq:#1}
                       \begin{eqnarray}
                       \label{eq:#1} }

\newcommand {\bearno}[1]{\mymarginpar{eq:#1}
                       \begin{eqnarray}
                       \nonumber}

\newcommand {\eear}{\end{eqnarray}}
\newcommand {\eearno}{\end{eqnarray}}
\newcommand {\bsel}{\left \{ \begin{array}{cl}}
\newcommand {\esel}{\end{array} \right.}

\newcommand {\bmat}[1]{\left [ \begin{array}{#1}}
\newcommand {\emat}{\end{array} \right ]}
\newcommand {\bsec}[2]{\mymarginpar{sec:#2}
                       \section{#1}
                       \label{sec:#2} }

\newcommand {\rsec}[1]{Section \ref{sec:#1}}

\newcommand {\bsubsec}[2]{\mymarginpar{sec:#2}
                       \subsection{#1}
                       \label{sec:#2} }

\def\R{I\kern-0.30em R}
\def\N{I\kern-0.30em N}
\def\P{I\kern-0.30em P}

\newcommand\RC{\mbox{RD}}
\newcommand\dbar{{\bar d}}
\newcommand\gbar{{\bar \gamma}}

\begin{document}

\title{A Mathematical Theory for Clustering in Metric Spaces}

\author{CHENG-SHANG CHANG
\affil{National Tsing Hua University}
WANJIUN LIAO
\affil{National Taiwan University}
YU-SHENG CHEN
\affil{National Taiwan University}
}

\author{Cheng-Shang~Chang,~\IEEEmembership{Fellow,~IEEE,}
        Wanjiun~Liao,~\IEEEmembership{Fellow,~IEEE,}
        Yu-Sheng~Chen,
        and~Li-Heng Liou
\thanks{C. S. Chang and L. H. Liou are with the Institute of Communications Engineering,
National Tsing Hua University,
Hsinchu 300, Taiwan, R.O.C.
email:  cschang@ee.nthu.edu.tw, dacapo1142@gmail.com}
\thanks{W. Liao and Y.-S. Chen are with Department of Electrical Engineering,
 National Taiwan University, Taipei, Taiwan, R.O.C.
email:  \{wjliao,r01921042\}@ntu.edu.tw.}
}

\maketitle
\thispagestyle{empty}

\begin{abstract}
Clustering is one of the most fundamental problems in data analysis and it has been  studied extensively in the literature. Though many clustering algorithms have been proposed,  clustering theories that justify the use of these clustering algorithms are still unsatisfactory.
In particular, one of the fundamental challenges  is to address the following question:
\begin{center}
What is a cluster in a set of data points?
\end{center}
In this paper, we  make an attempt to address such a question by considering a set of  data points associated with a distance measure (metric).
We first  propose a new {\em cohesion} measure in terms of the distance measure. Using the cohesion measure, we define a cluster as a set of points that are cohesive to themselves. For such a definition, we show there are various equivalent statements that have intuitive explanations.
We then consider the second question:
\begin{center}
How do we find clusters and good partitions of clusters under such a definition?
\end{center}
For such a question, we propose a hierarchical agglomerative  algorithm and a partitional algorithm. Unlike standard hierarchical agglomerative  algorithms,   our hierarchical agglomerative algorithm has a specific stopping criterion and it stops
with a partition of clusters. Our partitional algorithm, called the $K$-sets algorithm in the paper, appears to be a new iterative algorithm. Unlike the Lloyd iteration that needs two-step minimization, our $K$-sets algorithm only takes one-step minimization.

One of the most interesting findings of our paper is the duality result between a distance measure and a cohesion measure.  Such a duality result leads to a dual $K$-sets algorithm for clustering a set of data points with a cohesion measure.
The dual $K$-sets algorithm converges in the same way as a sequential version of the classical kernel $K$-means algorithm. The key difference is that a cohesion measure does not need to be positive semi-definite.

\end{abstract}




\bsec{Introduction}{introduction}


Clustering is one of the most fundamental problems in data analysis and it has a lot of applications in various fields, including Internet search for information retrieval, social network analysis for community detection, and  computation biology for clustering protein sequences.
The problem of clustering has been  studied extensively in the  literature (see e.g., the books \cite{theodoridispattern,rajaraman2012mining}
and the historical review papers \cite{jain1999data,jain2010data}). For a clustering problem, there is a set of data points (or objects) and a similarity
 (or dissimilarity) measure that measures how similar two data points are. The aim of a clustering algorithm is to cluster these data points so that data points within the same cluster are similar to each other and data points in different clusters are dissimilar.

As stated in \cite{jain2010data}, clustering algorithms can be divided into two groups: {\em hierarchical} and {\em partitional}. Hierarchical algorithms can further be divided into two subgroups: {\em agglomerative} and {\em divisive}. Agglomerative hierarchical algorithms, starting from each data point as a sole cluster, recursively merge two {\em similar}  clusters into a new cluster. On the other hand, divisive hierarchical algorithms, starting from the whole set as a single cluster, recursively divide a cluster into two {\em dissimilar} clusters. As such, there is a hierarchical structure of clusters from either a  hierarchical agglomerative clustering algorithm or a
 hierarchical divisive clustering algorithm.

Partitional algorithms do not have a hierarchical structure of clusters. Instead, they find all the clusters as a partition of the data points. The $K$-means algorithm is perhaps the simplest and the most widely used partitional algorithm for data points in an Euclidean space, where the Euclidean distance serves as the natural dissimilarity measure. The $K$-means algorithm starts from an initial partition of the data points into $K$ clusters. It then repeatedly carries out the Lloyd iteration \cite{lloyd1982least} that consists of the following two steps: (i) generate a new partition by assigning each data point to the closest cluster center, and (ii) compute the new cluster centers. The Lloyd iteration is known to reduce the sum of squared distance of each data point to its cluster center in each iteration and thus the $K$-means algorithm converges to a local minimum. The new cluster centers can be easily found if the data points are in a Euclidean space (or an inner product space). However, it is in general much more difficult to find the representative points for clusters, called medoids, if data points are  in a non-Euclidean space.
The refined $K$-means algorithms are commonly referred as the $K$-medoids algorithm (see e.g.,  \cite{kaufman2009finding,theodoridispattern,van2003new,park2009simple}). As the $K$-means algorithm (or the $K$-medoids algorithm) converges to a local optimum, it is quite sensitive to the initial choice of the partition. There are some recent works that provide various methods for selecting the initial partition that might lead to performance guarantees \cite{har2005fast,arthur2007k,bahmani2012scalable,jaiswal2012analysis,agarwal2013k}. Instead of using the Lloyd iteration to minimize the sum of squared distance of each data point to its cluster center, one can also formulate a clustering problem as an optimization problem with respect to a certain objective function and then solve the optimization problem by other methods. This then leads to kernel and spectral clustering methods (see e.g., \cite{shi2000normalized,ben2002support,ding2004k,camastra2005novel,chen2011parallel} and \cite{von2007tutorial,filippone2008survey} for reviews of the papers in this area). Solving the optimization problems formulated from the clustering problems are in general NP-hard and one has to resort to approximation algorithms \cite{balcan2013clustering}. In \cite{balcan2013clustering}, Balcan et al. introduced the concept of approximation stability that assumes all the partitions (clusterings) that have the objective values close to the optimum ones are close to the target partition. Under such an assumption, they proposed efficient  algorithms for clustering large data sets.

\begin{figure}[h]
\centering
\includegraphics[width=0.7\columnwidth]{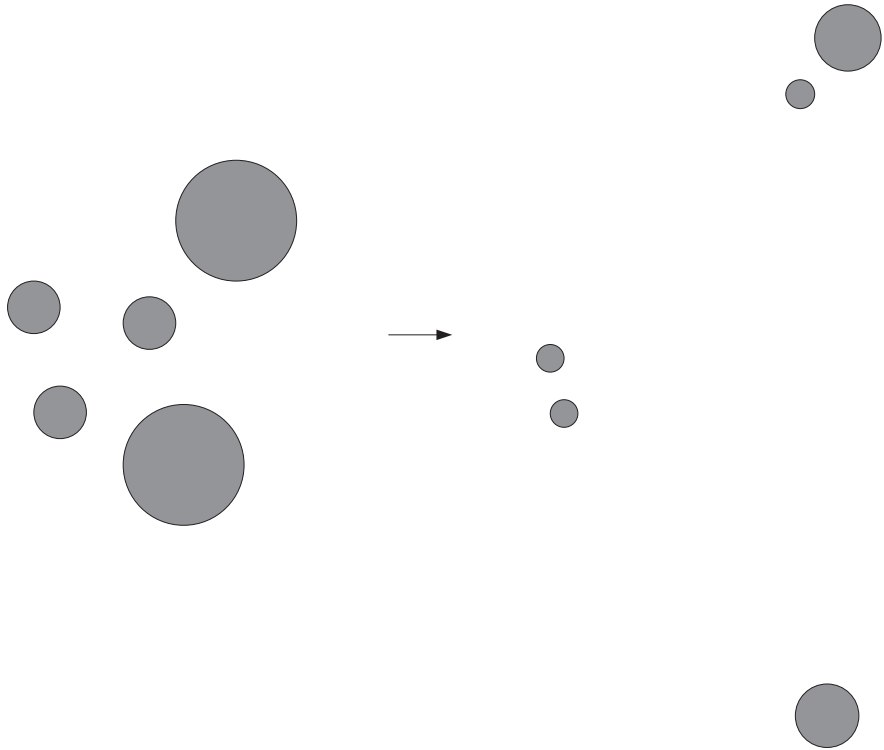}
\caption{A consistent change of 5 clusters.}
\label{fig:Figure1}
\end{figure}

Though there are already many clustering algorithms proposed in the literature,  clustering theories that justify the use of these clustering algorithms are still unsatisfactory.
As pointed out in \cite{von2005towards}, there are three commonly used approaches for developing a clustering theory: (i) an axiomatic approach that outlines a list of axioms for a clustering function (see e.g., \cite{jardine1971mathematical,wright1973formalization,puzicha2000theory,kleinberg2003impossibility,zadeh2009uniqueness,carlsson2014hierarchical}), (ii) an objective-based approach that provides a specific objective for a clustering function to optimize (see e.g., \cite{ben2009measures,balcan2013clustering}, and (iii) a definition-based approach that specifies the definition of clusters (see e.g, \cite{ester1996density,cuevas2001cluster,halkidi2008density}).   In  \cite{kleinberg2003impossibility}, Kleinberg adopted an axiomatic approach and showed  an impossibility theorem for finding a clustering function that satisfies the following three axioms:
\begin{description}
\item[(i)] Scale invariance: if we scale the dissimilarity measure by a constant factor, then the clustering function still outputs the same partition of clusters.
\item[(ii)] Richness:  for any specific partition of the data points, there exists a dissimilarity measure such that the clustering function outputs that partition.
\item[(iii)] Consistency: for a partition from the clustering function with respect to a specific dissimilarity measure, if we increase the dissimilarity measure between two points in different clusters and decrease the dissimilarity measure between two points in the same cluster, then the clustering function still outputs the same partition of clusters. Such a change of a dissimilarity measure is called a {\em consistent} change.
\end{description}
The impossibility theorem is based on the fundamental result that the output of any clustering function satisfying the scale invariance property and the consistency property is in a collection of {\em antichain} partitions, i.e., there is no partition in that collection that in turn is a {\em refinement} of another partition in that collection. As such, the richness property cannot be satisfied. In \cite{ben2009measures}, it was argued that the impossibility theorem is {\em not} an inherent feature of clustering. The key point in \cite{ben2009measures} is that the consistency property may not be a desirable property for a clustering function. This can be illustrated by considering a consistent change of 5 clusters in \rfig{Figure1}. The figure is redrawn from Figure 1 in \cite{ben2009measures} that originally consists of 6 clusters. On the left hand side of \rfig{Figure1}, it seems reasonable  to have a partition of 5 clusters. However, after the consistent change, a new partition of 3 clusters might be a better output than the original partition of 5 clusters. As such, they abandoned the three axioms for {\em clustering functions} and proposed another three similar axioms for Clustering-Quality Measures (CGM) (for measuring the quality of a partition). They showed the existence of a CGM that satisfies their three axioms for CGMs.

As for the definition-based approach, most of the definitions of a single cluster in the literature are  based on loosely defined terms \cite{theodoridispattern}. One exception is
 \cite{ester1996density}, where Ester et al. provided a precise definition of a single cluster based on the concept of  density-based reachability. A point $p$ is said to be directly density-reachable from another point $q$ if  point $p$ lies within the $\epsilon$-neighborhood of point $q$ and the $\epsilon$-neighborhood of point $q$ contains at least a minimum number of points.
  A point is said to be density-reachable from another point if they are connected by a sequence of directly density-reachable points. Based on the concept of density-reachability,
a cluster is defined as a maximal set of points that are density-reachable from each other. An intuitive way to see such a definition for a cluster in a set of data points is to convert the data set into a graph. Specifically, if we put a directed edge from one point $p$ to another point $q$ if point $p$ is directly density-reachable from point $q$, then a cluster simply corresponds to a {\em strongly connected component} in the graph.  One of the problems for such a definition is that it requires specifying two parameters,  $\epsilon$ and the minimum number of points in a $\epsilon$-neighborhood. As pointed out in \cite{ester1996density}, it is not an easy task to determine these two parameters.

In this paper, we make an attempt to develop a clustering theory in metric spaces.
In \rsec{cohesive}, we first address the question:
\begin{center}
What is a cluster in a set of data points in metric spaces?
\end{center}
For this, we first  propose a new {\em cohesion} measure in terms of the distance measure. Using the cohesion measure, we define a cluster as a set of points that are cohesive to themselves. For such a definition, we show in \rthe{clustereq} that there are various equivalent statements and these statements can be explained intuitively.
We then consider the second question:
\begin{center}
How do we find clusters and good partitions of clusters under such a definition?
\end{center}
For such a question, we propose a hierarchical agglomerative  algorithm in \rsec{halgorithms} and a partitional algorithm \rsec{triangularKsets}. Unlike standard hierarchical agglomerative  algorithms,   our hierarchical agglomerative algorithm has a specific stopping criterion. Moreover, we show in \rthe{community} that our hierarchical agglomerative algorithm returns a partition of clusters when it stops.
 Our partitional algorithm, called the $K$-sets algorithm in the paper, appears to be a new iterative algorithm. Unlike the Lloyd iteration that needs two-step minimization, our $K$-sets algorithm only takes one-step minimization. We further show in \rthe{triangular} that the $K$-sets algorithm
 converges in a finite number of iterations. Moreover, for $K=2$, the $K$-sets algorithm returns two clusters when the algorithm converges.

One of the most interesting findings of our paper is the duality result between a distance measure and a cohesion measure. In \rsec{cmeasures}, we first provide a general definition of a cohesion measure. We show that there is an induced distance measure, called the {\em dual distance measure}, for each cohesion measure. On the other hand, there is also an induced cohesion measure, called the {\em dual cohesion measure}, for each distance measure.
In \rthe{duality}, we further show that the dual distance measure of a dual cohesion measure of a distance measure is the distance measure itself. Such a duality result leads to a dual $K$-sets algorithm for clustering a set of data points with a cohesion measure.
The dual $K$-sets algorithm converges in the same way as a sequential version of the classical kernel $K$-means algorithm. The key difference is that a cohesion measure does not need to be positive semi-definite.


In Table \ref{tab:one}, we provide a list of notations used in this paper.

{
\tiny
\begin{table*}
\begin{center}
\caption{List of Notations\label{tab:one}}{%
\begin{tabular}{||l|l||}
\hline\hline
$\Omega=\{x_1, x_2, \ldots, x_n\}$ & The set of all data points \\
$n$ & The total number of data points \\
$d(x,y)$ & The distance between two points $x$ and $y$  \\
$\dbar(S_1, S_2)$
& The {\em average} distance between two sets $S_1$ and $S_2$ in \req{avg1111}\\
$\RC(x|| y)=d(x,y)-\dbar(\{x\}, \Omega)$  & The {\em relative} distance from $x$ to $y$ \\
$\RC(y)=\dbar(\Omega, \{y\})-\dbar(\Omega, \Omega)$ & The {\em relative} distance from a random point to $y$ \\
$\RC(S_1||S_2)=\dbar(S_1,S_2)-\dbar(S_1, \Omega)$ & The {\em relative distance} from a set $S_1$ to another set  $S_2$ \\
$\gamma(x,y)=\RC(y)-\RC(x||y)$ & The {\em cohesion measure} between two points $x$ and $y$ \\
$\gamma (S_1, S_2)=\sum_{x \in S_1}\sum_{y \in S_2}\gamma(x,y)$ & The {\em cohesion measure} between two sets $S_1$ and $S_2$ \\
$\Delta(x, S)=2\dbar(\{x\}, S)-\dbar(S,S)$ & The {\em triangular distance} from a point $x$ to a set $S$ \\
$Q=\sum_{k=1}^K \gamma(S_k,S_k)$ & The {\em modularity} for a partition $S_1, S_2, \ldots, S_K$ of $\Omega$ \\
$R=\sum_{k=1}^K \gamma(S_k,S_k)/|S_k|$ & The {\em normalized modularity} for a partition $S_1, S_2, \ldots, S_K$ of $\Omega$ \\ \hline
\hline
\end{tabular}}
\label{table:delay}
\end{center}
\end{table*}
}


\bsec{Clusters in metric spaces}{cohesive}

\bsubsec{What is a cluster?}{defcluster}

As pointed out in \cite{jain2010data}, one of the fundamental challenges associated with clustering is to address the following question:
\begin{center}
What is a cluster in a set of data points?
\end{center}

In this paper, we will develop a clustering theory that formally define a cluster for data points in a metric space.
 Specifically, we
 consider a set of $n$ data points, $\Omega=\{x_1, x_2, \ldots, x_n\}$ and a distance measure $d(x,y)$ for any two points $x$ and $y$ in $\Omega$.
The distance measure $d(\cdot, \cdot)$  is assumed to a {\em metric} and it satisfies
\begin{description}
\item[(D1)] $d(x,y) \ge 0$;
\item[(D2)] $d(x,x)=0$;
\item[(D3)] (Symmetric) $d(x,y)=d(y,x)$;
\item[(D4)] (Triangular inequality) $d(x,y) \le d(x,z)+d(z,y)$.
\end{description}
Such a metric assumption is stronger than the usual dissimilarity (similarity) measures \cite{LK03}, where the triangular inequality in general does not hold.
We also note that (D2) is usually stated as a necessary and sufficient condition in the literature, i.e., $d(x,y)=0$ if and only if $x=y$.
However, we only need the sufficient part in this paper.
Our approach begins with a definition-based approach. We first give
 a specific definition of what a cluster is (without the need of specifying any parameters) and show those axiom-like properties are indeed satisfied under our definition of a cluster.

\bsubsec{Relative distance and cohesion measure}{rdcm}

One important thing that we learn from the consistent change in \rfig{Figure1} is that a good partition of clusters should be looked at a global level and the {\em relative} distances among clusters should be considered as an important factor.
The distance measure between any two points only gives an absolute value and it does not tell us how close these two points are relative to the whole set of data points.  The key idea of defining the relative distance from one point $x$ to another point $y$  is to choose another random point $z$ as a reference point and compute the relative distance as the average of $d(x,y) -d(x,z)$ for all the points $z$ in $\Omega$.
This leads to the following definition of relative distance.

\bdefin{noravgxy} {\bf (Relative distance)}
The {\em relative distance} from a point $x$ to another point $y$, denoted by $\RC(x||y)$, is defined as follows:
\bear{noravg0000xy}
\RC(x|| y)&=& {1 \over n} \sum_{z \in \Omega}(d(x,y)-d(x,z))\nonumber\\
&=&d(x,y)-{1 \over n} \sum_{z \in \Omega}d(x,z).
\eear
The relative distance (from a random point) to a point $y$, denoted by $\RC(y)$, is defined as the average relative distance from a random point to $y$, i.e.,
\bear{noravg0000y}
\RC(y)&=& {1 \over n} \sum_{z \in \Omega} \RC(z||y)\nonumber\\
&=&{1 \over n}\sum_{z_2 \in \Omega} d(z_2,y)-{1 \over n^2} \sum_{z_2 \in \Omega}\sum_{z_1 \in \Omega}d(z_2,z_1).
\eear
\edefin

Note from \req{noravg0000xy} that in general $\RC(x || y)$ is not symmetric, i.e., $\RC(x || y) \ne \RC(y || x)$.
Also, $\RC(x||y)$ may not be {\em nonnegative}.
In the following, we extend the notion of relative distance from one point to another point to the relative distance from one set to another set.

\bdefin{noravg} {\bf (Relative distance)}
The {\em relative distance} from a set of points $S_1$ to another set of points $S_2$, denoted by $\RC(S_1||S_2)$, is defined as the average relative distance from a random point in $S_1$ to another random point in $S_2$,
 i.e.,
\beq{noravg0000}
\RC(S_1|| S_2)={1 \over {|S_1| \cdot |S_2|} }\sum_{x \in S_1} \sum_{y \in S_2} \RC(x || y).
\eeq
\edefin

Based on the notion of relative distance, we define a cohesion measure for two points $x$ and $y$ below.

\bdefin{cohesivexy}{\bf (Cohesion measure between two points)}
Define the {\em cohesion measure} between two points $x$ and $y$, denoted by $\gamma (x, y)$, as
the difference of the relative distance to $y$ and the relative distance from $x$ to $y$, i.e.,
\beq{coh1111xy}
\gamma(x, y) = \RC(y)-\RC(x||y).
\eeq
Two points $x$ and $y$ are said to be {\em cohesive} (resp. {\em incohesive}) if $\gamma(x, y) \ge 0$ (resp. $\gamma(x, y) \le 0$).
\edefin

In view of \req{coh1111xy}, two points $x$ and $y$ are cohesive if the relative distance from $x$ to $y$ is not larger than the relative distance (from a random point) to $y$.

Note from \req{noravg0000xy} and \req{noravg0000y} that
\bear{coh2222xy}
\gamma(x, y) &=&\RC(y)-\RC(x||y) \nonumber \\
& =&{1 \over n}\sum_{z_2 \in \Omega} d(z_2,y)+{1 \over n} \sum_{z_1 \in \Omega}d(x,z_1)\nonumber \\
&&\quad\quad-{1 \over n^2} \sum_{z_2 \in \Omega}\sum_{z_1 \in \Omega}d(z_2,z_1)-d(x,y)  \label{eq:coh2222xya}\\
&=&{1 \over n^2} \sum_{z_2 \in \Omega}\sum_{z_1 \in \Omega}\Big (d(x,z_1)+d(z_2,y)\nonumber \\
&&\quad\quad-d(z_1,z_2)-d(x,y) \Big)
 \label{eq:coh2222xyb}.
\eear


Though there are many ways to define a cohesion measure for a set of data points in a metric space, our definition of the cohesion measure in \rdef{cohesivexy} has the following
 four desirable properties.
Its proof is based on the representations in \req{coh2222xya} and \req{coh2222xyb} and it is given in Appendix A.

\bprop{cohesivexy}
\begin{description}
\item[(i)] (Symmetry) The cohesion measure is symmetric, i.e., $\gamma(x,y)=\gamma(y,x)$.
\item[(ii)] (Self-cohesiveness) Every data point is cohesive to itself, i.e., $\gamma(x,x) \ge 0$.
\item[(iii)] (Self-centredness) Every data point is more cohesive to itself than to another point, i.e., $\gamma(x,x) \ge \gamma(x,y)$ for all $y \in \Omega$.
\item[(iv)] (Zero-sum) The sum of the cohesion measures  between a data point to all the points in $\Omega$ is zero, i.e., $\sum_{y \in \Omega}\gamma(x,y)=0$.
\end{description}
\eprop

These four properties can be understood intuitively by viewing a cohesion measure between two points as a ``binding force'' between those two points.
The symmetric property ensures that the binding force is reciprocated.  The self-cohesiveness property ensures that each point is self-binding. The self-centredness property further ensures that the self binding force is always stronger than the binding force to the other points.
In view of the zero-sum property, we know for every point $x$ there are points that are incohesive to $x$ and each of these points has a negative binding force to $x$. Also, there are points that are cohesive to $x$ (including $x$ itself from the self-cohesiveness property) and each of these points has a positive force to $x$. As such, the binding force will naturally ``push'' points into ``clusters.''

To further understand the intuition of the cohesion measure, we can think of $z_1$ and $z_2$ in \req{coh2222xyb} as two random points
that are used as reference points.
Then two points $x$ and $y$ are cohesive if $d(x,z_1)+d(z_2,y) \ge d(z_1,z_2)+d(x,y)$ for two reference points $z_1$ and $z_2$ that are randomly chosen from $\Omega$.
In \rfig{cohesive}, we show an illustrating example for such an intuition in ${\cal R}^2$. In \rfig{cohesive}(a), point $x$ is close to one reference point $z_1$ and point $y$ is close to
the other reference point $z_2$. As such, $d(x,z_1)+d(z_2,y) \le d(z_1,z_2)$ and thus these two points $x$ and $y$ are incohesive. In \rfig{cohesive}(b),  point $x$ is not that close to $z_1$ and point $y$ is not that close to
 $z_2$. However, $x$ and $y$ are on the two opposite sides of the segment between the two reference points $z_1$ and $z_2$.
As such, there are two triangles in this graph: the first triangle consists of the three points $x, z_1$, and $w$, and
the second triangle consists of the three points $y, z_2$, and $w$. From the triangular inequality, we then have
$d(w,z_1)+d(x,w) \ge d(x,z_1)$ and $d(y,w)+d(w,z_2) \ge d(y,z_2)$. Since $d(w,z_1)+d(w,z_2)=d(z_1, z_2)$ and
$d(x,w)+d(y,w)=d(x,y)$, it then follows that $d(z_1, z_2)+d(x,y) \ge d(x, z_1) + d(y, z_2)$. Thus, points $x$ and $y$ are also incohesive in \rfig{cohesive}(b).
In \rfig{cohesive}(c),  point $x$ is not that close to $z_1$ and point $y$ is not that close to
 $z_2$ as in \rfig{cohesive}(b). Now $x$ and $y$ are on the same side of the segment between the two reference points $z_1$ and $z_2$. There are two triangles in this graph: the first triangle consists of the three points $x, y$, and $w$, and
the second triangle consists of the three points $z_1, z_2$, and $w$. From the triangular inequality, we then have
$d(x,w)+d(w,y) \ge d(x,y)$ and $d(w,z_1)+d(w,z_2) \ge  d(z_1, z_2)$.
Since $d(x,w)+d(w,z_1)=d(x, z_1)$ and
$d(w,y)+d(w, z_2)=d(z_2, y)$, it then follows that $d(x, z_1) + d(y, z_2) \ge d(z_1, z_2)+d(x,y)$. Thus, points $x$ and $y$ are cohesive in \rfig{cohesive}(c). In view of \rfig{cohesive}(c), it is intuitive to see that two points $x$ and $y$ are cohesive if
{\em they both are far away from the two reference points and they both are close to each other}.

\bfig{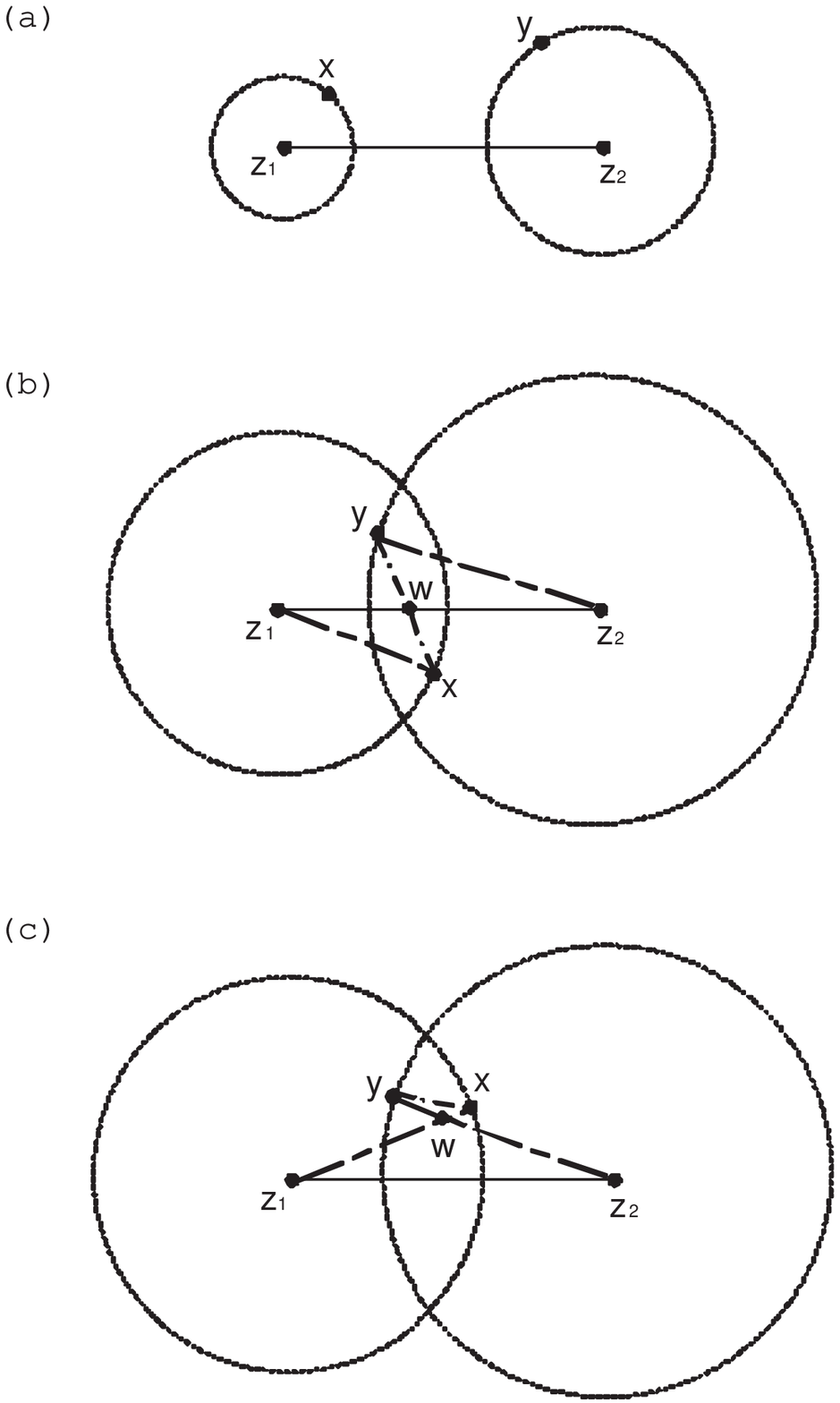}{2.5 in} \efig{cohesive}{Illustrating examples of the cohesion measure in ${\cal R}^2$: (a) incohesive as $d(x, z_1)+d(y,z_2) \le d(z_1, z_2)$, (b) incohesive as $d(w,z_1)+d(x,w) \ge d(x,z_1)$ and $d(y,w)+d(w,z_2) \ge d(y,z_2)$, and (c) cohesive $d(x,w)+d(w,y) \ge d(x,y)$ and $d(w,z_1)+d(w,z_2) \ge  d(z_1, z_2)$.}

The notions of relative distance and cohesion measure are also related to the notion of relative centrality in our previous work \cite{chang2013relative}.
To see this, suppose that we sample two points $x$ and $y$ from $\Omega$ according to the following bivariate distribution:
\beq{bivar1111}
p(x,y)={{e^{-\theta d(x,y)}} \over {\sum_{u \in \Omega}\sum_{v \in \Omega}e^{-\theta d(u,v)}}}, \quad \theta >0.
\eeq
Let $P_X(x)=\sum_{y \in \Omega}p(x,y)$ and $P_Y(y)=\sum_{x \in \Omega}p(x,y)$ be the two marginal distributions.
Then  one can verify that the covariance
$p(x,y)-P_X(x)P_Y(y)$ is proportional to the cohesion measure $\gamma(x,y)$ when $\theta \downarrow 0$.
Intuitively, two points $x$ and $y$ are cohesive if they are positively correlated according to the sampling in \req{bivar1111} when $\theta$ is very small.

Now we extend the cohesion measure between two points to the cohesion measure between two sets.

\bdefin{cohesive}{\bf (Cohesion measure between two sets)}
Define the {\em cohesion measure} between two sets $S_1$ and $S_2$, denoted by $\gamma (S_1, S_2)$, as the sum of the cohesion measures of all the pairs of two points (with one point in $S_1$ and the other point in $S_2$), i.e.,
\beq{coh1111}
\gamma(S_1, S_2) = \sum_{x \in S_1}\sum_{y \in S_2}\gamma(x,y).
\eeq
Two sets $S_1$ and $S_2$ are said to be {\em cohesive} (resp. {\em incohesive}) if $\gamma(S_1, S_2) \ge 0$ (resp. $\gamma(S_1, S_2) \le 0$).
\edefin

\bsubsec{Equivalent statements of clusters}{clusters}

Now we define what a cluster is in terms of the cohesion measure.

\bdefin{cluster} {\bf (Cluster)}
A nonempty set $S$ is called a {\em cluster} if it is cohesive to itself, i.e.,
\beq{cluster1111}
\gamma(S, S) \ge 0.
\eeq
\edefin

In the following, we show the first main theorem of the paper. Its proof is given in Appendix B.

\bthe{clustereq}
Consider a {\em nonempty} set
 $S$ that is not equal to $\Omega$. Let $S^c = \Omega \backslash S$ be the set of points that are not in $S$.
  Also, let $\dbar (S_1, S_2)$ be  the average distance between two randomly selected points with one point in $S_1$ and another point in  $S_2$, i.e.,
\beq{avg1111}
\dbar(S_1, S_2) = {1 \over {|S_1| \times {|S_2|}}} \sum_{x \in S_1}\sum_{y \in S_2} d(x,y).
\eeq
The following statements are equivalent.
\begin{description}
\item[(i)] The set $S$ is a cluster, i.e., $\gamma(S,S)\ge 0$.
\item[(ii)] The set $S^c$ is a cluster, i.e., $\gamma (S^c, S^c) \ge 0$.
\item[(iii)] The two sets $S$ and $S^c$ are incohesive, i.e., $\gamma(S, S^c) \le 0$.
\item[(iv)] The set  $S$ is more cohesive to itself than to $S^c$, i.e., $\gamma(S,S) \ge \gamma(S,S^c)$.
\item[(v)] $2\dbar (S, \Omega)-\dbar(\Omega, \Omega)-\dbar (S,S)\ge 0$.
\item[(vi)] The relative distance from $\Omega$ to $S$ is not smaller than the relative distance from $S$ to $S$, i.e., $\RC(\Omega||S) \ge \RC(S||S)$.
\item[(vii)]  The relative distance from $S^c$ to $S$ is not smaller than the relative distance from $S$ to $S$, i.e., $\RC(S^c||S) \ge \RC(S||S)$.
\item[(viii)] $2\dbar(S, S^c)- \dbar(S,S)-\dbar(S^c,S^c) \ge 0$.
\item[(ix)] The relative distance from $S$ to $S^c$ is not smaller than the relative distance from $\Omega$ to $S^c$, i.e., $\RC(S||S^c) \ge \RC(\Omega|| S^c)$.
    \item[(x)]   The relative distance from $S^c$ to $S$  is not smaller than the relative distance from $\Omega$ to $S$, i.e., $\RC(S^c||S) \ge \RC(\Omega||S)$.
\end{description}
\ethe

One surprise finding in \rthe{clustereq}(ii) is that the set $S^c$ is also a cluster.
This shows that the points inside $S$ are cohesive and the points outside $S$ are also cohesive. Thus, there seems a boundary between $S$ and $S^c$ from the cohesion measure. Another surprise finding is in \rthe{clustereq}(viii).
 One usually would expect that a cluster $S$ should satisfy $\dbar(S,S) \le \dbar(S,S^c)$. But it seems our definition of a cluster is much weaker than that.
Regarding the scale invariance property, it is easy to see from \rthe{clustereq}(viii) that the inequality there is still satisfied if we scale the distance measure by a constant factor. Thus, a cluster of data points is still a cluster after scaling the distance measure by a constant factor. Regarding the richness property, we argue that there exists a distance measure such that any subset of points in $\Omega$ is a cluster. To see this, we simple let the distance between any two points in the subset be equal to 0 and the distance between a point outside the subset to a point in the subset be equal to 1. Since a point $x$ itself is a cluster, i.e., $\gamma(x,x)\ge 0$, we then have $\gamma(x,y)=\gamma(x,x)\ge 0$ for any two points $x$ and $y$ in the subset. From \req{cluster1111}, the subset is a cluster under such a choice of the distance measure.  Furthermore, one can also see from \rthe{clustereq}(vii) that for a cluster $S$,
if we  decrease the relative distance between two points in $S$ and increase the relative distance between one  point in $S$ and another point in $S^c$, then the set $S$ is still a cluster under such a "consistent" change.

We also  note that in our proof of \rthe{clustereq} we only need $d(\cdot,\cdot)$ to be symmetric. As such, the results in \rthe{clustereq} also hold even when the triangular inequality is not satisfied.


\bsec{A hierarchical agglomerative algorithm}{halgorithms}

Once we define what a cluster is, our next question is
\begin{center}
How do we find clusters and good partitions of clusters?
\end{center}

For this, we turn to an objective-based approach. We will show that clusters can be found by optimizing two specific objective functions by a  hierarchical algorithm in \rsec{halgorithms} and a partitional algorithm in \rsec{triangularKsets}.

 In the following, we first define a quality measure for
a partition of $\Omega$.

\bdefin{index}{\bf (Modularity)}
Let $S_k$, $k=1,2, \ldots, K$, be a partition of $\Omega=\{x_1,x_2,\ldots, x_n\}$, i.e.,
$S_k \cap S_{k^\prime}$ is an empty set for $k \ne k^\prime$ and $\cup_{k=1}^K S_k=\Omega$.
The  modularity index $Q$ with respect to the partition $S_k$, $k=1,2, \ldots, K$, is
defined as follows:
\beq{index1111}
Q=\sum_{k=1}^K \gamma (S_k,S_k) .
\eeq
\edefin

\begin{algorithm}[t]
\KwIn{A data set $\Omega=\{x_1, x_2, \ldots, x_n\}$ and a distance measure $d(\cdot, \cdot)$.
}
\KwOut{A partition of clusters $\{S_1, S_2, \ldots, S_K\}$.}
 Initially, $K=n$; $S_i=\{x_i\}$, $i=1,2,\ldots, n$\;
Compute the cohesion measures $\gamma(S_i, S_j)=\gamma(x_i,x_j)$ for all $i, j=1,2, \ldots, n$\;
  \While{there exists some $i$ and $j$ with $\gamma(S_i, S_j)>0$}{
   Merge $S_i$ and $S_j$ into a new set $S_k$, i.e., $S_k =S_i \cup S_j$\;
   $\gamma(S_k, S_k)= \gamma(S_i, S_i)+2 \gamma (S_i, S_j) + \gamma (S_j, S_j)$\;
   \For{each $\ell \ne k$}{
       $\gamma (S_k, S_\ell) =\gamma(S_\ell, S_k)=\gamma(S_i, S_\ell)+\gamma(S_j, S_\ell)$\;
   }
   $K=K-1$;
}
  Reindex the $K$ remaining sets to $\{S_1, S_2, \ldots, S_K\}$\;
\caption{The Hierarchical Agglomerative Algorithm}
\label{alg:hierarchical}
\end{algorithm}

Based on such a quality measure, we can thus formulate the clustering problem as an optimization problem for finding a partition $S_1, S_2, \ldots, S_K$ (for some unknown $K$) that maximizes the modularity index $Q$.
Note that
\bear{index2222}
&&Q=\sum_{k=1}^K \gamma (S_k,S_k)=\sum_{k=1}^K \sum_{x \in S_k} \sum_{y \in S_k}\gamma(x,y) \nonumber\\
&&=\sum_{x \in \Omega}\sum_{y \in \Omega} \gamma(x,y) \delta_{c(x), c(y)},
\eear
where $c(x)$ is the cluster of $x$ and $\delta_{c(x), c(y)}=1$ if $x$ and $y$ are in the same cluster.
In view of \req{index2222}, another way to look at the optimization problem is to find the assignment of each point to a cluster.
However, it was shown in \cite{Brandes08} that finding the optimal
assignment for modularity maximization is NP-complete in the strong sense and thus heuristic algorithms, such as hierarchical algorithms and partitional algorithms are commonly used in the literature for solving the modularity maximization problem.

In Algorithm \ref{alg:hierarchical}, we propose a hierarchical agglomerative clustering algorithm that converges to a local optimum of this objective. The algorithm starts from $n$ clusters with each point itself as a cluster. It then recursively merges two disjoint cohesive clusters to form a new cluster until either there is  a single cluster left or all the remaining clusters are incohesive. There are two main differences between a standard hierarchical agglomerative clustering algorithm and ours:
\begin{description}
\item[(i)] Stopping criterion: in a standard hierarchical agglomerative clustering algorithm, such as single linkage or complete linkage, there is no stopping criterion. Here our algorithm stops  when all the remaining clusters are incohesive.
\item[(ii)] Greedy selection: our algorithm only needs to select a pair of cohesive clusters to merge. It does not need to be the most cohesive pair. This could potentially speed up the algorithm in a large data set.
\end{description}

In the following theorem, we show  that the modularity index $Q$ in \req{index1111} is non-decreasing in every iteration
of the hierarchical agglomerative
 clustering algorithm and it indeed produces clusters. Its proof is given in Appendix C.

\bthe{community}
\begin{description}
\item[(i)] Every set returned by the hierarchical agglomerative clustering algorithm  is indeed
a cluster.
\item[(ii)] For the hierarchical agglomerative clustering algorithm,
the modularity index is non-decreasing in every iteration and thus converges to a local optimum.
\end{description}
\ethe

 As commented before,  our algorithm only requires to find a pair of cohesive clusters to merge in each iteration. This is different from the greedy selection
in \cite{theodoridispattern}, Chapter 13, and \cite{Newman04}.
Certainly, our hierarchical agglomerative clustering algorithm can also be operated in a greedy manner. As in \cite{Newman04}, in each iteration we can merge the two clusters that result in the largest increase of the modularity index, i.e., the most cohesive pair.
It is well-known (see e.g., the book \cite{rajaraman2012mining}) that a n\"aive implementation of a greedy hierarchical agglomerative clustering algorithm
has $O(n^3)$  computational complexity and the computational complexity can be further reduced to $O(n^2 \log(n))$ if priority queues are implemented for the greedy selection. We also note that there are several hierarchical agglomerative clustering algorithms proposed in the literature for community detection in networks (see e.g., \cite{NG04,blondel2008fast,Lambiotte2010,Delvenne2010}).  These algorithms are also based on ``modularity'' maximization. Among them, the fast unfolding algorithm in \cite{blondel2008fast} is the fast one as there is a second phase of building a new (and much smaller) network whose nodes are the communities found during the previous phase. The Newman and Girvan modularity in \cite{NG04} is based on a probability measure from a random selection of an edge in a network (see \cite{chang2013relative} for more detailed discussions) and this is different from the distance metric used in this paper.

\bfig{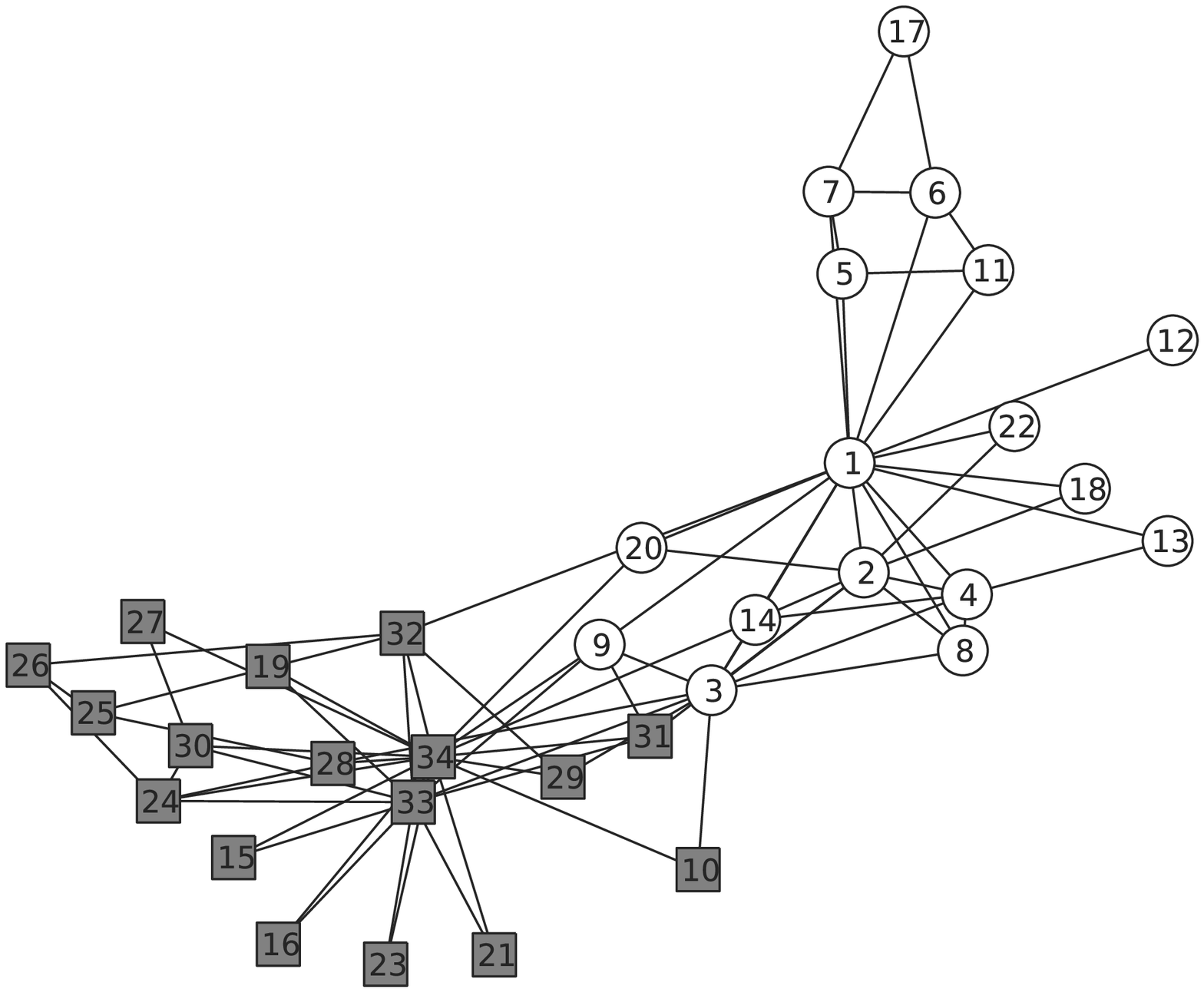}{3.0 in} \efig{Zachary}{The network of friendships between individuals in the
karate club study of Zachary\cite{Zachary77}. The instructor  and the administrator
are represented by nodes 1 and 34, respectively.
Squares represent individuals who ended up with the administrator  and circles represent those who
ended up with the instructor.}

In the following, we provide an illustrating example for our  hierarchical agglomerative clustering  algorithm
by using the greedy selection of the most cohesive pair.

\bex{zachary}{\bf (Zachary's karate club)}{\em As in \cite{NG04,Chang:11:AGP}, we consider the Zachary's karate club friendship network \cite{Zachary77} in \rfig{Zachary}. The set of data was
observed by Wayne Zachary \cite{Zachary77} over the course of two years
in the early 1970s at an American university. During the course
of the study, the club split into two groups because of a dispute
between its administrator (node 34 in \rfig{Zachary}) and its instructor (node 1 in \rfig{Zachary}).

\bfig{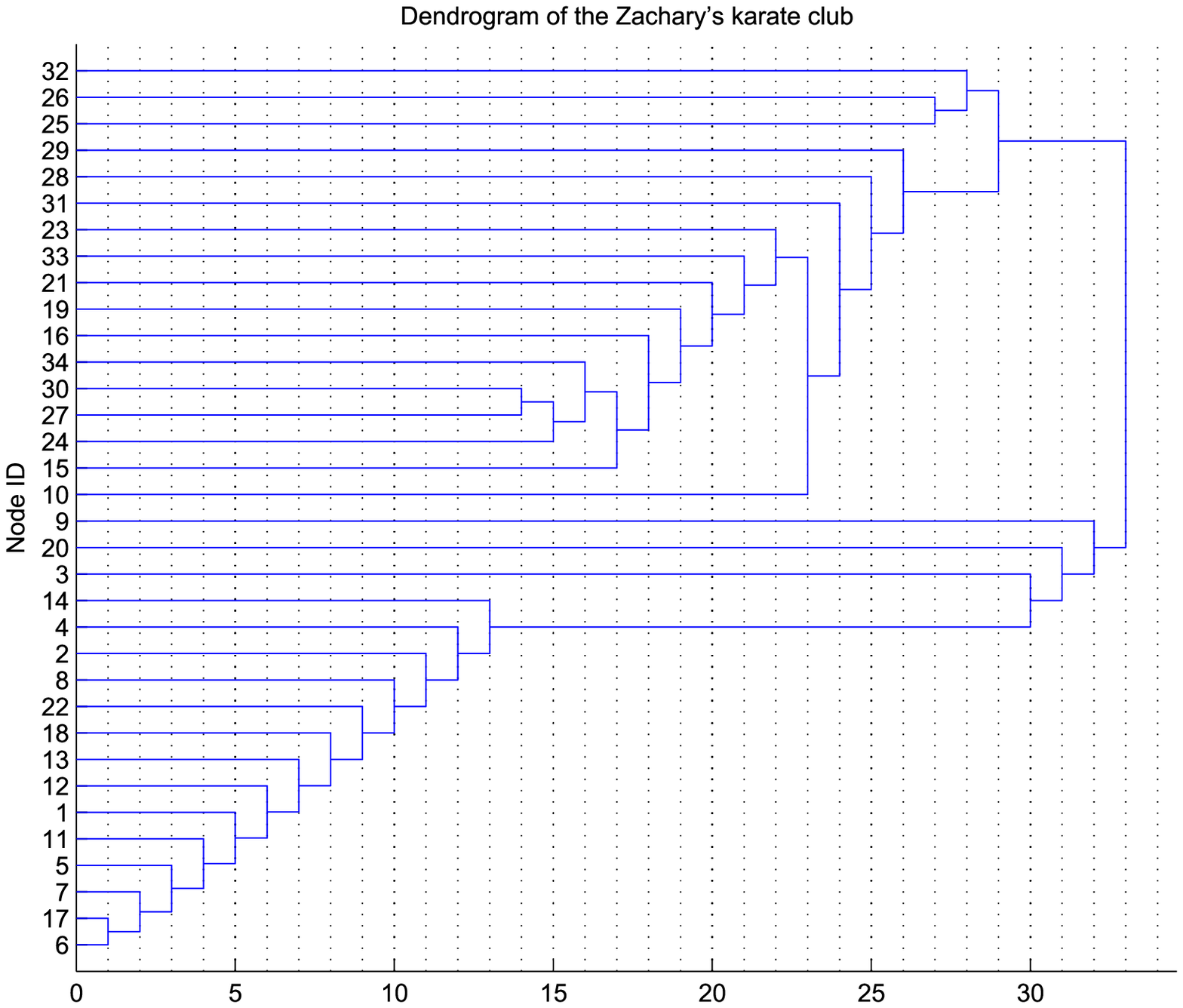}{3.5 in} \efig{dendrogram}{The dendrogram from our greedy hierarchical agglomerative clustering algorithm for the Zachary karate club friendship network.}

In Figure 6, we show the dendrogram generated by using our  hierarchical agglomerative clustering  algorithm
with the greedy selection of the most cohesive pair in each iteration.
The distance measure is the geodesic distance of the graph in \rfig{Zachary}.  The algorithm stops when there are three incohesive clusters left, one led by the administrator (node 34), one led by the instructor (node 1), and person number 9 himself.
According to \cite{Zachary77}, there was an interesting story for person number 9. He was a weak supporter for the administrator. However, he was only three weeks away from a test for black belt (master status) when the split of the club occurred. He would have had to give up his rank if he had joined the administrator's club. He ended up with the instructor's club.
We also run an additional step for our algorithm (to merge the pair with the largest cohesive measure) even though the remaining three clusters are incohesive. The
additional step reveals that person number 9 is clustered into the instructor's club.
}
\eex

\bsec{A partitional algorithm}{triangularKsets}

\bsubsec{Triangular distance}{triangular}

In this section, we consider another objective function.

\bdefin{indexn}{\bf (normalized modularity)}
Let $S_k$, $k=1,2, \ldots, K$, be a partition of $\Omega=\{x_1,x_2,\ldots, x_n\}$, i.e.,
$S_k \cap S_{k^\prime}$ is an empty set for $k \ne k^\prime$ and $\cup_{k=1}^K S_k=\Omega$.
The  normalized modularity index $R$ with respect to the partition $S_k$, $k=1,2, \ldots, K$, is
defined as follows:
\beq{totalstr1111}
R=\sum_{k=1}^K {1 \over {|S_k|}}\gamma(S_k,S_k).
\eeq
\edefin

Unlike the hierarchical agglomerative clustering algorithm in the previous section, in this section we assume that $K$ is fixed and known in advance. As such, we may use an approach similar to the classical $K$-means algorithm by iteratively assigning each point to the nearest set (until it converges). Such an approach requires a measure that can measure how close a point $x$ to a set $S$ is. In the $K$-means algorithm, such a measure is defined as the square of the distance between $x$ and the centroid of $S$. However, there is no centroid for a set in a non-Euclidean space and we need to come up with another measure.

\bfig{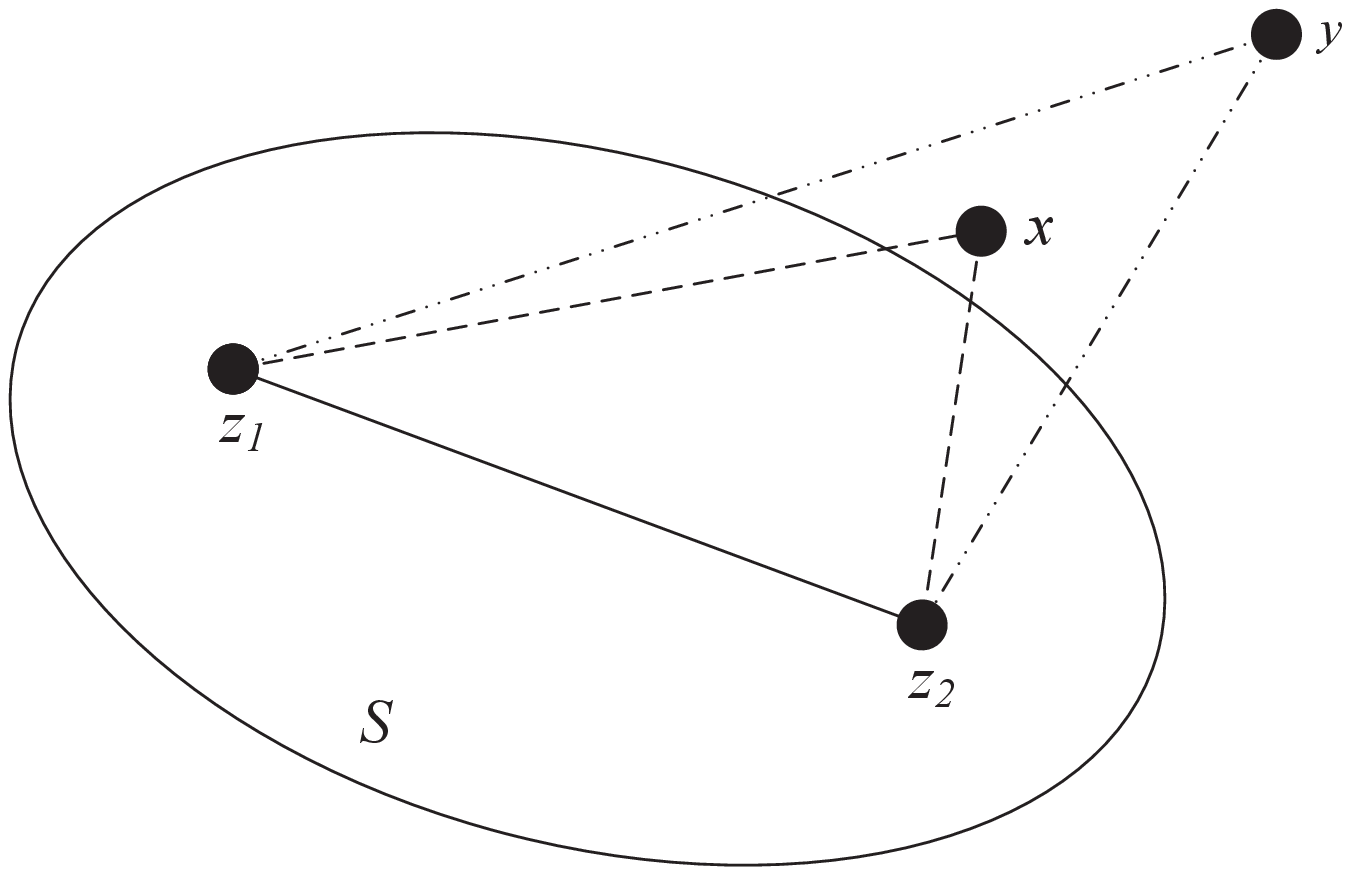}{3.0 in} \efig{triangular}{An illustration of the triangular distance in ${\cal R}^2$.}

Our idea for measuring the distance from a point $x$ to a set $S$ is to randomly choose two points $z_1$ and $z_2$ from $S$ and consider the three sides of the triangle $x$, $z_1$ and $z_2$.  Note that the triangular inequality guarantees that $d(x,z_1)+d(x,z_2)-d(z_1, z_2) \ge 0$.
Moreover, if $x$ is close to $z_1$ and $z_2$, then $d(x,z_1)+d(x,z_2)-d(z_1, z_2)$ should also be small. We illustrate such an intuition in \rfig{triangular}, where there are two points $x$ and $y$ and a set $S$ in ${\cal R}^2$.
Such an intuition leads to the following definition of triangular distance from a point $x$ to a set $S$.

\bdefin{triangular}{\bf (Triangular distance)} The {\em triangular distance} from a point $x$ to a set $S$, denoted by $\Delta(x, S)$, is defined as follows:
\beq{triang1111}
\Delta(x, S)= {1 \over {|S|^2}}\sum_{z_1 \in S} \sum_{z_2 \in S}\Big (d(x,z_1)+d(x,z_2)-d(z_1, z_2)\Big ).
\eeq
\edefin

In the following lemma, we show several properties of the triangular distance and its proof is given in  Appendix D.


\blem{totalstr}
\begin{description}
\item[(i)]  \beq{triang1111b} \Delta(x, S)=2\dbar(\{x\}, S) - \dbar(S,S) \ge 0.\eeq
\item[(ii)]
\beq{totalstr2222}
\Delta(x, S)= \gamma(x,x)-{2 \over |S|} \gamma(\{x\},S)+ {1 \over {|S|^2}}\gamma(S,S).
\eeq
\item[(iii)] Let $S_k$, $k=1,2, \ldots, K$, be a partition of $\Omega=\{x_1,x_2,\ldots, x_n\}$. Then
\beq{totalstr3333}
\sum_{k=1}^K \sum_{x \in S_k}\Delta(x, S_k)=\sum_{x \in \Omega}\gamma(x,x)-R.
\eeq
\item[(iv)] Let $S_k$, $k=1,2, \ldots, K$, be a partition of $\Omega=\{x_1,x_2,\ldots, x_n\}$ and $c(x)$ be the index of the set to which $x$ belongs, i.e., $x \in S_{c(x)}$. Then
\bear{totalstr3355}
\sum_{k=1}^K \sum_{x \in S_k}\Delta(x, S_k)&=&\sum_{k=1}^K \sum_{x \in S_k}\dbar(\{x\}, S_k)\nonumber\\ &=&\sum_{x \in \Omega}\dbar (\{x\}, S_{c(x)}).
\eear
\end{description}
\elem

The first property of this lemma is to  represent triangular distance by the average distance.
The second property is to represent the triangular distance by the cohesion measure. Such a property plays an important role for the duality result in \rsec{cmeasures}. The third property shows that the optimization problem for maximizing the normalized modularity $R$ is equivalent to the optimization problem that minimizes the sum of the triangular distance of each point to its set. The fourth property further shows that such an optimization problem is also equivalent to the optimization problem that minimizes the sum of the average distance of each point to its set.
 Note that $\dbar(\{x\}, S_k)={1 \over {|S_k|}}\sum_{y \in S_k}d(x,y)$. The objective for maximizing the normalized modularity $R$ is also equivalent to minimize
$$\sum_{k=1}^K {1 \over {|S_k|}} \sum_{x \in S_k}\sum_{y \in S_k}d(x,y).$$
This is different from the $K$-median objective, the $K$-means objective and the min-sum objective addressed in \cite{balcan2013clustering}.

\bsubsec{The $K$-sets algorithm}{Ksets}

In the following, we propose a partitional clustering algorithm, called the $K$-sets algorithm in Algorithm \ref{alg:ksets},  based on the triangular distance. The algorithm is very simple. It starts from an arbitrary partition of the data points that contains $K$ disjoint sets.
Then for each data point, we assign the data point to the closest set in terms of the triangular distance. We repeat the process until there is no further change. Unlike the Lloyd iteration that needs two-step minimization, the $K$-sets algorithm only takes one-step minimization. This might give the $K$-sets  algorithm the computational advantage over the $K$-medoids algorithms \cite{kaufman2009finding,theodoridispattern,van2003new,park2009simple}.

\begin{algorithm}[t]
\KwIn{A data set $\Omega=\{x_1, x_2, \ldots, x_n\}$, a distance measure $d(\cdot,\cdot)$,  and the number of sets $K$.
}
\KwOut{A partition of sets $\{S_1, S_2, \ldots, S_K\}$.}

\noindent {\bf (0)} Initially, choose arbitrarily $K$ disjoint nonempty sets $S_1 , \ldots, S_K$ as a partition of $\Omega$.

\noindent {\bf (1)} \For{$i=1, 2, \ldots, n$}{

\noindent
Compute the triangular distance $\Delta(x_i, S_k)$ for each set $S_k$ by using \req{triang1111b}.\\
Find the set to which the point $x_i$ is closest in terms of the triangular distance. \\
Assign point $x_i$ to that set.}

\noindent {\bf (2)} Repeat from (1) until there is no further change.
\caption{The K-sets Algorithm}
\label{alg:ksets}
\end{algorithm}

In the following theorem, we show the convergence of the $K$-sets algorithm. Moreover, for $K=2$, the $K$-sets algorithm yields two clusters.  Its proof is given in Appendix E.

\bthe{triangular}
\begin{description}
\item[(i)] In the $K$-sets algorithm based on the triangular distance, the normalized modularity is increasing when there is a change, i.e.,  a point is moved from one set to another. Thus, the algorithm converges to a local optimum of the normalized modularity.
\item[(ii)] Let $S_1, S_2, \ldots, S_K$ be the $K$ sets when the algorithm converges. Then for all $i \ne j$,  the   two sets $S_i$ and $S_j$ are two clusters if these two sets are viewed in isolation (by removing the data points not in $S_i \cup S_j$ from $\Omega$).
\end{description}
\ethe

An immediate consequence of \rthe{triangular} (ii) is that for $K=2$, the two sets $S_1$ and $S_2$ are clusters when the algorithm converges.
However, we are not able to show that for $K \ge 3$ the $K$ sets, $S_1, S_2, \ldots, S_K$, are clusters in $\Omega$.
On the other hand, we are not able to find a counterexample either. All the numerical examples that we have tested for $K \ge 3$ yield $K$ clusters.

\bsubsec{Experiments}{exp}

In this section, we report several experimental results for the $K$-sets algorithm: including the dataset with two rings in Section \ref{sec:tworings}, the stochastic block model in Section \ref{sec:sbm}, and
the mixed National Institute of Standards and Technology dataset in Section \ref{sec:MNIST}.

\subsubsection{Two rings}
\label{sec:tworings}

\bfig{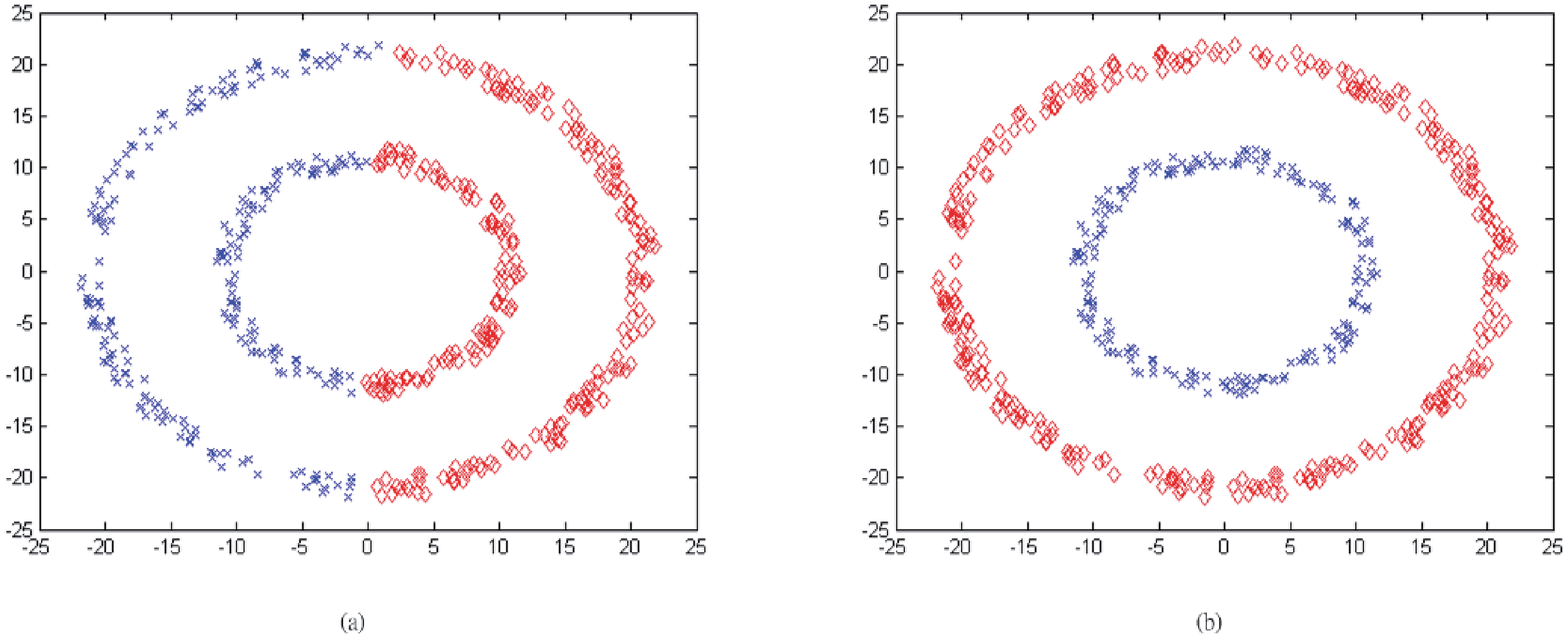}{3.0 in} \efig{tworing}{Two rings: (a) a clustering result by the $K$-means algorithm, and (b) a clustering result by the $K$-sets algorithm.}

In this section, we first provide an illustrating example for the $K$-sets algorithm.

In \rfig{tworing}, we generate two rings by randomly placing 500 points in ${\cal R}^2$. The outer (resp. inner) ring consists of 300 (resp. 200) points. The radius of a point in the outer (resp. inner) ring is uniformly distributed between 20 and 22 (resp. 10 and 12). The angle of each point is uniformly distributed between 0 and $2 \pi$.
In \rfig{tworing}(a), we show a typical clustering result by using the classical $K$-means algorithm with $K=2$. As the centroids of these two rings are very close to each other, it is well-known that the $K$-means algorithm does not perform well for the two rings.  Instead of using the Euclidean distance as the distance measure for our $K$-sets algorithm, we first convert the two rings into a graph by adding an edge between two points with  the Euclidean distance less than 5. Then the distance measure between two points is defined as the geodesic distance of these two points in the graph. By doing so, we can then easily separate these rings by using the $K$-sets algorithm with $K=2$ as shown in \rfig{tworing}(b).

 The purpose of this example is to show the limitation of the applicability of the $K$-means algorithm. The data points for the
  $K$-means algorithm need to be in some Euclidean space. On the other hand, the data points for the $K$-sets algorithms only need to be in some metric space.  As such, the distance matrix constructed from a graph cannot be directly applied by the $K$-means algorithm while it is still applicable for the $K$-sets algorithm.

\subsubsection{Stochastic block model}
\label{sec:sbm}

The stochastic block model (SBM), as a generalization of the Erd\"os-R\'enyi random graph \cite{erdds1959random}, is a commonly used method for generating random graphs that can be used for benchmarking community detection algorithms \cite{saade2014spectral,SBM-generator}. In a stochastic block model with $q$ blocks (communities), the total number of nodes in the random graph are evenly distributed to these $q$ blocks. The probability that there is an edge between two nodes within the same block is $p_{in}$ and the probability that there is an edge between two nodes in two different blocks is $p_{out}$. These edges are generated independently. Let $c_{in} = n \cdot p_{in}$ , $c_{out} = n \cdot p_{out}$. Then it is known \cite{saade2014spectral} that these $q$ communities can be detected (in theory for a large network) if
\beq{transition}
 |c_{in} - c_{out}| > q \sqrt{\text{mean degree}}.
\eeq

  In this paper, we use MODE-NET \cite{SBM-generator} to run SBM. Specifically, we consider a stochastic block model with two blocks. The number of nodes in the stochastic block model is 1,000 with 500 nodes in each of these two blocks. The
 average degree of a node is set to be 3.
  The values of $c_{in}-c_{out}$ of these graphs are in the range from 2.5 to 5.9 with a common step of 0.1. We generate 20 graphs for each $c_{in}-c_{out}$. Isolated vertices are removed. Thus, the exact numbers of vertices used in this experiment are slightly less than than 1,000.

We compare our $K$-sets algorithm with some other community detection algorithms, such as OSLOM2 \cite{lancichinetti2011finding}, infomap \cite{rosvall2007maps,rosvall2010map}, and fast unfolding \cite{blondel2008fast}. The metric used for the $K$-sets algorithm for each sample of the random graph is the resistance distance, and this is pre-computed by NumPy \cite{van2011numpy}. The resistance distance matrix (denoted by $R=(R_{i,j})$) can be derived from the pseudo inverse of the adjacency matrix (denoted by $\Gamma=(\Gamma_{i,j})$)  as follows: \cite{bapat2003simple}:
\[ R_{i,j} =
  \begin{cases}
    0,       & \quad \text{if } i = j,\\
    \Gamma_{i,i} + \Gamma_{j,j} - \Gamma_{i,j} - \Gamma_{j,i}, & \quad \text{otherwise}.\\
  \end{cases}
\]
The $K$-sets algorithm and OSLOM2 are implemented in C++, and the others are all taken from igraph \cite{igraph} and are implemented in C with python wrappers. In Table \ref{tab:runtime}, we show the average running times for these four algorithms  over 700 trials. The pre-computation time for the $K$-sets algorithm is the time to compute the distance matrix.  Except infomap, the other three algorithms are very fast. In \rfig{sbm}, we  compute the normalized mutual information measure (NMI) by using a built-in function in igraph \cite{igraph} for the results obtained from these four algorithms.
Each point is averaged over 20 random graphs from the stochastic block model. The error bars are the 95\% confidence intervals.
In this stochastic block model, the theoretical phase transition threshold from \req{transition} is $c_{in}-c_{out}=3.46$. It seems that the $K$-sets algorithm is able to detect these two blocks when $c_{in}-c_{out} \ge 4.5$. Its performance in that range is better than  infomap \cite{rosvall2007maps,rosvall2010map}, fast unfolding \cite{blondel2008fast} and OSLOM2 \cite{lancichinetti2011finding}. We note that the comparison is not exactly fair as the other three algorithms do not have the information of the number of blocks (communities).

\begin{table}[ht]
\centering
\caption{Average running time (in seconds).\label{tab:runtime}}
\begin{tabular}{ccccc}
\hline
                & infomap  & fast unfolding & OSLOM2  & $K$-sets    \\ \hline
pre-computation & 0        & 0              & 0       & 2.3096    \\ \hline
running         & 0.7634   & 0.0074         & 0.0059  & 0.0060    \\ \hline
total           & 0.7634   & 0.0074         & 0.0059  & 2.3156    \\ \hline
\end{tabular}
\end{table}

\bfig{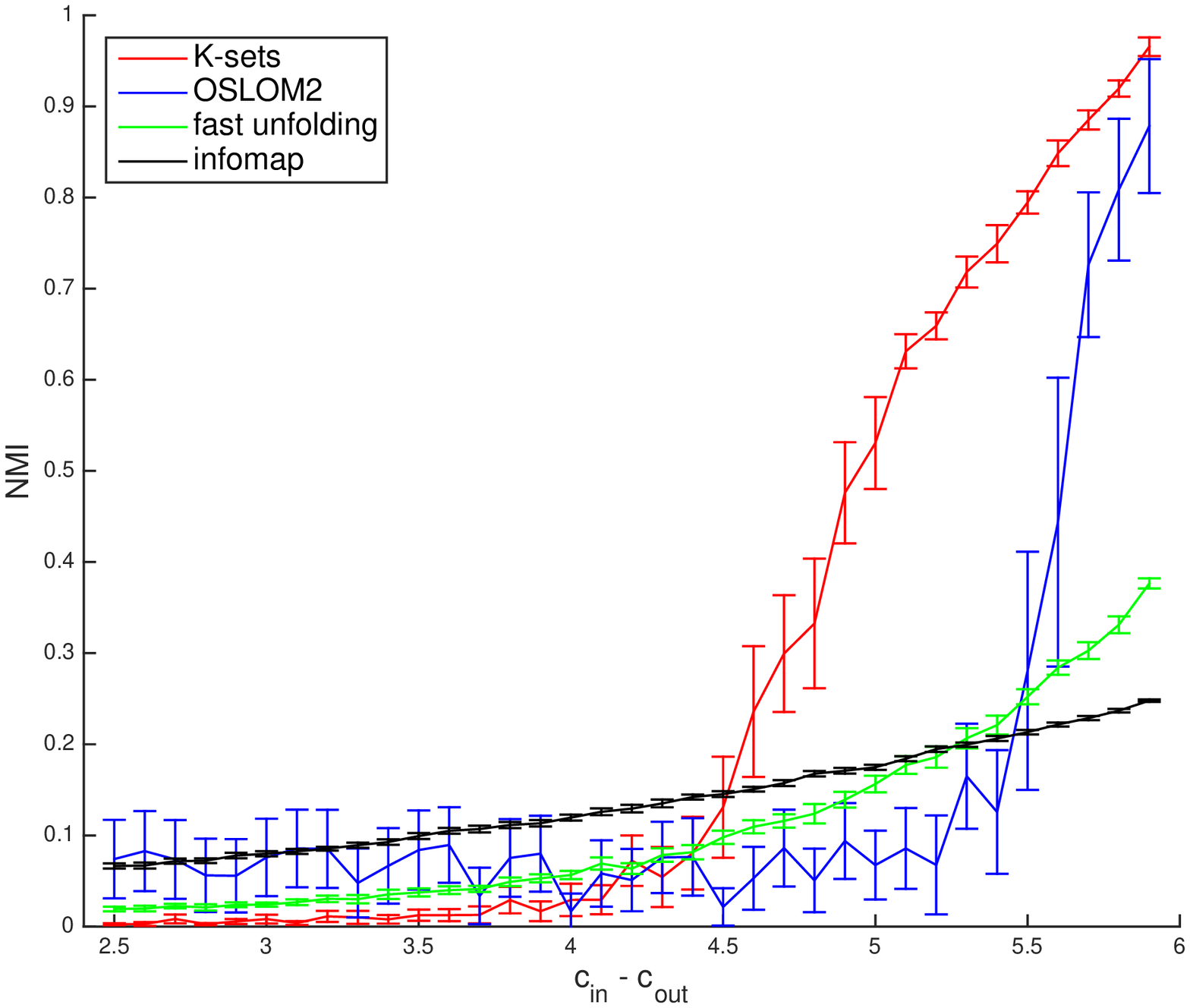}{3.0 in} \efig{sbm}{Comparison of infomap \cite{rosvall2007maps,rosvall2010map}, fast unfolding \cite{blondel2008fast}, OSLOM2 \cite{lancichinetti2011finding} and $K$-sets for the stochastic block model with two blocks. Each point is averaged over 20 such graphs. The error bars are the 95\% confidence intervals. The theoretical phase transition threshold in this case is 3.46.}

\subsubsection{Mixed National Institute of Standards and Technology dataset}
\label{sec:MNIST}

In this section, we consider a real-world dataset, the mixed National Institute of Standards and Technology dataset (the MNIST dataset) \cite{lecun-mnisthandwrittendigit-2010}.
The MNIST dataset contains 60,000 samples of hand-written digits. These samples are 28$\times$28 pixels grayscale images (i.e., each of the image is a 784 dimensional data point). For our experiments, we select the first 1,000 samples from each set of the digit 0 to 9 to create a total number of 10,000 samples.

To fairly evaluate the performance of the $K$-sets algorithm, we compare the $K$-sets algorithm with two  clustering algorithms in which the number of clusters is also known a priori, i.e., the $K$-means++ algorithm \cite{scikit-learn} and the $K$-medoids algorithm \cite{beauchamp2011msmbuilder2}. For the MNIST dataset, the number of clusters is 10 (for the ten digits, $0,1,2,\ldots, 9$). The $K$-means++ algorithm is an improvement of the standard $K$-means algorithm with a specific method to choose the initial centroids of the $K$ clusters. Like the $K$-sets algorithm, the $K$-medoids algorithm is also a clustering algorithm that uses a distance measure. The key difference between the $K$-sets algorithm and the $K$-medoids algorithm is that we use the triangular distance to a set for the assignment of each data point and the $K$-medoids algorithm uses the distance to a medoid for such an assignment.  The Euclidean distance between two data points (samples from the MNIST dataset) for the $K$-medoids algorithm and the $K$-sets algorithm are pre-computed by NumPy \cite{van2011numpy}. The $K$-sets algorithm is implemented in C++, and the others are implemented in C with python wrappers.
All the programs are executed on an Acer Altos-T350-F2 machine with two Intel(R) Xeon(R) CPU E5-2690 v2 processors. In order to have a fair comparison of their running times, the parallelization of each program is disabled, i.e., only one core is used in these experiments. We assume that the input data is already stored in the main memory and the time consumed for I/O is  not recorded.

In Table \ref{tab:runtime2}, we show the average running times for these three algorithms  over 100 trials.
Both the $K$-medoids algorithm and the $K$-sets algorithm need to compute the distance matrix and this is shown in the row marked with the pre-computation time. The total running times for these three algorithms are roughly the same for this experiment.
In \rfig{mnist}, we  compute the normalized mutual information measure (NMI) by using a built-in function in igraph \cite{igraph} for the results obtained from these three algorithms.
Each point is averaged over 100 trials. The error bars are the 95\% confidence intervals.
In view of  \rfig{mnist}, the $K$-sets algorithm outperforms the $K$-means++ algorithm  and the $K$-medoids algorithm for the MNIST dataset.
One possible explanation for this is that both the the $K$-means++ algorithm  and the $K$-medoids algorithm only select a single representative data point for a cluster and that representative data point may not be able to represent the whole cluster well enough. On the other hand, the $K$-sets algorithm uses the triangular distance that takes the distance to every point in a cluster into account.

\begin{table}[ht]
\centering
\caption{Average running time (in seconds). \label{tab:runtime2}}
\begin{tabular}{cccc}
\hline
                & $K$-means++ & $K$-medoids & $K$-sets \\ \hline
pre-computation & 0         & 36.981    & 36.981 \\ \hline
running         & 49.940    & 1.228     & 1.801  \\ \hline
total           & 49.940    & 38.209    & 38.782 \\ \hline
\end{tabular}
\end{table}

\bfig{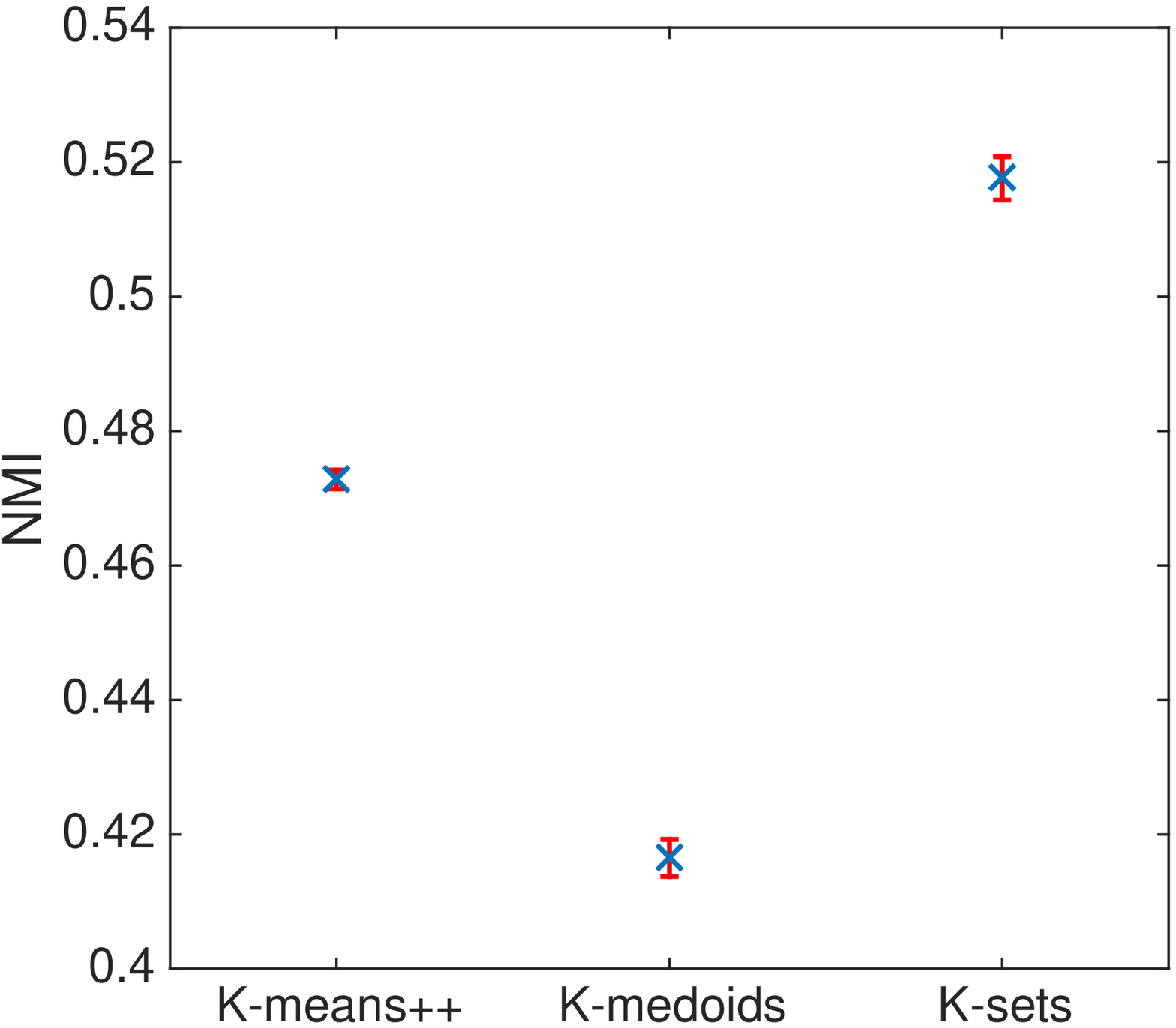}{2.5 in} \efig{mnist}{Comparison of $K$-means++ \cite{scikit-learn},  $K$-medoids \cite{beauchamp2011msmbuilder2} and $K$-sets for the MNIST dataset. Each point is averaged over 100 trials. The error bars are the 95\% confidence intervals.}

\bsec{Duality between  a cohesion measure and a distance measure}{cmeasures}

\bsubsec{The duality theorem}{dtheorem}

In this section, we show the duality result between a cohesion measure and a distance measure.
In the following, we first provide a {\em general} definition for a cohesion measure.

\bdefin{cmeasures}
A measure between two points $x$ and $y$, denoted by $\beta(x,y)$, is called a {\em cohesion measure} for a set of data points $\Omega$ if
it satisfies the following three properties:
\begin{description}
\item[(C1)] (Symmetry) $\beta(x,y)=\beta(y,x)$ for all $x, y \in \Omega$.
\item[(C2)] (Zero-sum) For all $x \in \Omega$, $\sum_{y \in \Omega}\beta (x,y)=0$.
\item[(C3)] (Triangular inequality) For all $x, y, z$ in $\Omega$,
\beq{cmeas1111}
\beta(x,x)+\beta (y,z) -\beta (x,z)-\beta (x, y) \ge 0.
\eeq
\end{description}
\edefin

In the following lemma,  we show that the specific cohesion measure defined in \rsec{cohesive} indeed
satisfies (C1)--(C3) in \rdef{cmeasures}. Its proof is given in Appendix F.

\blem{csimdist}
Suppose that $d(\cdot,\cdot)$ is a distance measure for $\Omega$, i.e., $d(\cdot,\cdot)$ satisfies (D1)--(D4).
Let
\bear{csim7777}
\beta(x,y)&=&{1 \over n}\sum_{z_2 \in \Omega} d(z_2,y)+{1 \over n} \sum_{z_1 \in \Omega}d(x,z_1)\nonumber \\&&\quad-{1 \over n^2} \sum_{z_2 \in \Omega}\sum_{z_1 \in \Omega}d(z_2,z_1)-d(x,y).
\eear
Then $\beta(x,y)$ is a cohesion measure for $\Omega$.
\elem

We know from \req{coh2222xya} that the cohesion measure $\gamma(\cdot,\cdot)$ defined in \rsec{cohesive} has the following representation:
\bearn
&&\gamma(x,y)={1 \over n}\sum_{z_2 \in \Omega} d(z_2,y)+{1 \over n} \sum_{z_1 \in \Omega}d(x,z_1)\\
&&\quad-{1 \over n^2} \sum_{z_2 \in \Omega}\sum_{z_1 \in \Omega}d(z_2,z_1)-d(x,y).
\eearn
As a result of \rlem{csimdist}, it also satisfies (C1)--(C3) in  \rdef{cmeasures}. As such, we call the cohesion measure $\gamma(\cdot,\cdot)$ defined in \rsec{cohesive}
the {\em dual cohesion measure} of the distance measure $d(\cdot,\cdot)$.

On the other hand, if $\beta(x,y)$ is a cohesion measure for $\Omega$, then there is an induced distance measure and it can be viewed as the {\em dual distance measure} of the cohesion measure  $\beta(x,y)$.
This is shown in the following lemma and its proof is given in Appendix G.

\blem{cinduced}
Suppose that $\beta(\cdot, \cdot)$ is a cohesion measure for $\Omega$. Let
\beq{cind1111}
d(x,y)=(\beta(x,x)+\beta(y,y))/2 -\beta(x,y).
\eeq
Then $d(\cdot,\cdot)$ is a distance measure that satisfies (D1)--(D4).
\elem

In the following theorem, we show the duality result. Its proof is given in Appendix H.

\bthe{duality}
Consider a set of data points $\Omega$.
For a  distance measure $d(\cdot,\cdot)$ that satisfies (D1)--(D4), let
\bear{dual1111}
&&d^*(x,y)={1 \over n}\sum_{z_2 \in \Omega} d(z_2,y)+{1 \over n} \sum_{z_1 \in \Omega}d(x,z_1)\nonumber\\
&&\quad-{1 \over n^2} \sum_{z_2 \in \Omega}\sum_{z_1 \in \Omega}d(z_2,z_1)-d(x,y)
\eear
be the dual cohesion measure of $d(\cdot,\cdot)$. On the other hand, For a cohesion measure
$\beta(\cdot,\cdot)$ that satisfies (C1)--(C3), let
\beq{dual2222}
\beta^*(x,y)=(\beta(x,x)+\beta(y,y))/2 -\beta(x,y)
\eeq
be the dual distance measure of $\beta(\cdot,\cdot)$.
Then $d^{**}(x,y)=d(x,y)$ and $\beta^{**}(x,y)=\beta(x,y)$ for all $x, y \in \Omega$.
\ethe

\bsubsec{The dual $K$-sets algorithm}{dual}

For the $K$-sets algorithm, we need to have a distance measure.
In view of the duality theorem between a cohesion measure and a distance measure,
 we propose the dual $K$-sets algorithm in Algorithm \ref{alg:dualksets} that uses a cohesion measure.
As before, for a cohesion measure $\gamma(\cdot,\cdot)$ between two points, we define the cohesion measure between two sets $S_1$ and $S_2$ as
\beq{dualu1111}
\gamma(S_1, S_2)=\sum_{x \in S_1} \sum_{y \in S_2}\gamma(x,y).
\eeq
Also, note from \req{totalstr2222} that the triangular distance from a point $x$ to a set $S$ can be computed
by using the cohesion measure as follows:
\beq{dualu2255}
\Delta(x, S)= \gamma(x,x)-{2 \over {|S|}}\gamma(\{x\},S)+ {1 \over {|S|^2}}\gamma(S,S).
\eeq

\begin{algorithm}[t]
\KwIn{A data set $\Omega=\{x_1, x_2, \ldots, x_n\}$, a cohesion measure $\gamma(\cdot, \cdot)$,  and the number of sets $K$.
}
\KwOut{A partition of sets $\{S_1, S_2, \ldots, S_K\}$.}

\noindent {\bf (0)} Initially, choose arbitrarily $K$ disjoint nonempty sets $S_1 , \ldots, S_K$ as a partition of $\Omega$.

\noindent {\bf (1)} \For{$i=1, 2, \ldots, n$}{

\noindent
Compute the triangular distance $\Delta(x_i, S_k)$ for each set $S_k$ by using \req{dualu2255}.\\
Find the set to which the point $x_i$ is closest in terms of the triangular distance. \\
Assign point $x_i$ to that set.}

\noindent {\bf (2)} Repeat from (1) until there is no further change.
\caption{The dual K-sets Algorithm}
\label{alg:dualksets}
\end{algorithm}

As a direct result of the duality theorem in \rthe{duality} and the convergence result of the $K$-sets algorithm in \rthe{triangular}, we have the following convergence result for the dual $K$-sets algorithm.

\bcor{triangularb}
As in \req{totalstr1111}, we define the normalized modularity as $\sum_{k=1}^K {1 \over {|S_k|}}\gamma(S_k,S_k)$.
For the dual $K$-sets algorithm, the normalized modularity is increasing when there is a change, i.e.,  a point is moved from one set to another. Thus, the algorithm converges to a local optimum of the normalized modularity. Moreover, for $K=2$, the dual $K$-sets algorithm yields two clusters when the algorithm converges.
\ecor

\bsubsec{Connections to the kernel $K$-means algorithm}{kernel}

In this section, we show the connection between the dual $K$-sets algorithm and the kernel $K$-means algorithm in the literature (see e.g., \cite{Dhillon04b}).
Let us consider the $n \times n$ matrix $\Gamma=(\gamma_{i,j})$ with
$\gamma_{i,j}=\gamma(x_i, x_j)$ being the cohesion measure between $x_i$ and $x_j$. Call the matrix $\Gamma$ the cohesion matrix (corresponding to the cohesion measure $\gamma(\cdot,\cdot)$). Since $\gamma(x_i, x_j)=\gamma(x_j, x_i)$, the matrix $\Gamma$ is symmetric and thus have real eigenvalues $\lambda_k, k=1,2, \ldots, n$. Let $\bf I$ be the $n \times n$ identity matrix and $\sigma \ge -\min_{1 \le k \le n}{\lambda_k}$. Then the matrix $\tilde \Gamma =\sigma {\bf I} +\Gamma$ is positive semi-definite
as its $n$ eigenvalues $\tilde \lambda_k = \sigma +\lambda_k$, $k=1,2, \ldots, N$ are all nonnegative. Let $v_k=(v_{k,1}, v_{k,2}, \ldots, v_{k,n})^T$, $k=1,2,\ldots, n$ be the eigenvector of $\Gamma$ corresponding to the eigenvalue $\lambda_k$. Then $v_k$ is also the
 eigenvector of $\tilde \Gamma$ corresponding to the eigenvalue $\tilde \lambda_k$. Thus, we can decompose the matrix $\tilde \Gamma$ as follows:
\beq{kernel1111}
\tilde \Gamma =\sum_{k=1}^n \tilde \lambda_k v_k v_k^T,
\eeq
where $v_k^T$ is the transpose of $v_k$. Now we choose the mapping $\phi :\Omega \mapsto {\cal R}^n$ as follows:
\beq{kernel2222}
\phi(x_i) =\Big (\sqrt{\tilde \lambda_1}v_{1,i}, \sqrt{\tilde \lambda_2}v_{2,i}, \ldots, \sqrt{\tilde \lambda_n}v_{n,i} \Big )^T,
\eeq
for $i=1,2 \ldots, n$.
Note that
\bear{kernel2233}
\phi(x_i)^T \cdot \phi(x_j) &=&\sum_{k=1}^n \tilde \lambda_k v_{k,i} v_{k,j}\nonumber\\
&=& (\tilde \Gamma)_{i,j}
= \sigma  \delta_{i,j}+\gamma(x_i, x_j),
\eear
where $\delta_{i,j}=1$ if $i=j$ and 0 otherwise.

The ``centroid'' of a set $S$ can be represented by the corresponding centroid in ${\cal R}^n$, i.e.,
\beq{kernel3333}
{1 \over |S|}\sum_{y \in S}\phi(y),
\eeq
and the square of the ``distance'' between a point $x$ and the ``centroid'' of a set $S$ is
\bear{kernel4444}
&&\Big (\phi(x)-{1 \over {|S|}}\sum_{y \in S}\phi(y) \Big )^T\cdot \Big (\phi(x)-{1 \over {|S|}}\sum_{y \in S}\phi(y)\Big ) \nonumber \\
&&=\phi(x)^T \cdot \phi(x) -{2 \over {|S|}}\sum_{y \in S}\phi(x)^T\cdot \phi(y)
\nonumber\\
&&\quad\quad
+{1 \over {|S|^2}}\sum_{y_1 \in S}\sum_{y_2 \in S}\phi(y_1)^T\cdot \phi(y_2) \nonumber\\
&&=\gamma(x,x) +\sigma  -{2 \over {|S|}}\sum_{y \in S}\gamma(x,y)-{{2 \sigma} \over {|S|}} 1_{\{x \in S\}}\nonumber\\
&&\quad\quad
+{1 \over {|S|^2}}\sum_{y_1 \in S}\sum_{y_2 \in S}\gamma(y_1, y_2)+{{|S| \sigma} \over {|S|^2}}\nonumber \\
&&=(1-{{2 } \over {|S|}} 1_{\{x \in S\}}+{1 \over {|S|}})\sigma +\gamma(x,x)-{2 \over {|S|}}\gamma(\{x\}, S)\nonumber\\
&&\quad\quad
+{1 \over {|S|^2}}\gamma(S,S),
\eear
where $1_{\{x \in S\}}$ is the indicator function that has value 1 if $x$ is in $S$ and 0 otherwise.
In view of \req{totalstr2222}, we then have
 \bear{kernel4455}
&&\Big (\phi(x)-{1 \over {|S|}}\sum_{y \in S}\phi(y) \Big )^T\cdot \Big (\phi(x)-{1 \over {|S|}}\sum_{y \in S}\phi(y)\Big )\nonumber\\
&&=(1-{{2 } \over {|S|}} 1_{\{x \in S\}}+{1 \over {|S|}})\sigma +\Delta(x, S),
\eear
where $\Delta(x, S)$ is the triangular distance from a point $x$ to a set $S$.
 Thus, the square of the ``distance'' between a point $x$ and the ``centroid'' of a set $S$ is $(1-{1 \over {|S|}})\sigma+ \Delta(x, S)$ for a point $x \in S$ and
$(1+{1 \over {|S|}})\sigma+ \Delta(x, S)$ for a point $x \not \in S$. In particular, when $\sigma=0$,
the dual $K$-sets algorithm  is the same as the sequential kernel $K$-means algorithm for the kernel $\tilde \Gamma$.
Unfortunately, the matrix $\tilde \Gamma$ may not be positive semi-definite if $\sigma$ is chosen to be 0. As indicated in \cite{Dhillon04b}, a large $\sigma$ decreases (resp. increases)
the distance from a point $x$ to a set $S$ that contains (resp. does not contain) that point. As such, a point is more unlikely to move from one set to another set and the kernel $K$-means algorithm is thus more likely to be trapped in a local optimum.

To summarize,
the dual $K$-sets algorithm operates in the same way as a sequential version of the classical kernel $K$-means algorithm  by viewing the matrix $\Gamma$ as a kernel. However, there are two key differences between the dual $K$-sets algorithm and the classical kernel $K$-means algorithm: (i) the dual $K$-sets algorithm guarantees the convergence even though the matrix $\Gamma$ from a cohesion measure
is not positive semi-definite, and (ii) the dual $K$-sets algorithm can only be operated {\em sequentially} and the kernel $K$-means algorithm can be operated in batches.
To further illustrate the difference between these two algorithms,
 we show in the following two examples that a cohesion  matrix  may not be positive semi-definite and a positive semi-definite matrix may not be a cohesion matrix.

\bex{Counter1}{}
{\em
In this example, we show there is a cohesion  matrix $\Gamma$ that is not a positive semi-definite matrix.
\beq{counter1111}
{\Gamma}=\left (
\begin{array}{rrrrr}
0.44   & 0.04  &    0.04  &  0.04 &   -0.56 \\
0.04   & 0.64  &   -0.36  & -0.36 &    0.04 \\
0.04   &-0.36  &    0.64  & -0.36 &    0.04 \\
0.04   &-0.36  &   -0.36  &  0.64 &    0.04 \\
-0.56  & 0.04  &    0.04  &  0.04 &    0.44
\end{array}
\right ) .
\eeq
The eigenvalues of this matrix are $-0.2$, $0$, $1$, $1$, and $1$.
}
\eex

\bex{Counter2}{}
{\em
In this example, we show there is a positive semi-definite matrix $M=(m_{i,j})$ that is not an cohesion matrix.
\beq{counter2222}
M=\left (
\begin{array}{rrrr}
0.375  & -0.025 &   -0.325 &   -0.025 \\
-0.025 & 0.875  & -0.025   &   -0.825 \\
-0.325 & -0.025 &    0.375 &   -0.025 \\
-0.025 & -0.825 &   -0.025 &    0.875
\end{array}
\right ) .
\eeq
The eigenvalues of this matrix are $0$, $0.1$, $0.7$, and $1.7$.
Even though the matrix $M$ is symmetric and has all its row sums and column sums being 0,
it is still not a cohesion matrix as $m_{1,1}-m_{1,2}-m_{1,4}+m_{2,4}=-0.4<0$.
}
\eex

\bsubsec{Constructing a cohesion measure from a similarity measure}{construction}

A similarity measure is in general defined as a bivariate function of two {\em distinct} data points and it is often characterized
by a square matrix without specifying the diagonal elements.
In the following, we show how one can construct a cohesion measure from a symmetric bivariate function by further specifying the
diagonal elements. Its proof is given in Appendix I.

\bprop{csim}
Suppose  a  bivariate function $\beta_0 : \Omega\times \Omega \mapsto {\cal R}$ is symmetric, i.e.,
$\beta_0(x,y)=\beta_0(y,x)$.
Let $\beta_1(x,y)=\beta_0(x,y)$ for all $x \ne y$
and specify $\beta_1(x,x)$ such that
\beq{csim1111}
\beta_1(x,x)\ge \max_{x \ne y \ne z} [\beta_1 (x,z)+\beta_1 (x, y)-\beta_1 (y,z)].
\eeq
Also, let
\bear{csim2222}
&&\beta(x,y)=\beta_1(x,y)- {1 \over n}\sum_{z_1 \in \Omega}\beta_1(z_1,y) \nonumber
\\
&&\quad -{1 \over n}\sum_{z_2 \in \Omega}\beta_1(x,z_2) +{1 \over {n^2}}\sum_{z_1 \in \Omega}\sum_{z_2 \in \Omega}\beta_1(z_1,z_2).
\eear
Then $\beta(x,y)$ is a cohesion measure for $\Omega$.
\eprop

We note that one simple choice for specifying $\beta_1(x,x)$ in \req{csim1111} is to set
\beq{csim4488}
\beta_1(x,x)=2 \beta_{\max} - \beta_{\min},
\eeq
where
\beq{csim4490}
\beta_{\max} =\max_{x \ne y}\beta(x,y),
\eeq
and
\beq{csim4492}
\beta_{\min}=\min_{x \ne y}\beta(x,y).
\eeq
In particular, if the similarity measure $\beta(x,y)$ only has values 0 and 1 as in the adjacency matrix of a simple undirected graph, then one can simply choose $\beta_1(x,x)=2$ for all $x$.

\bex{largenet2}{\bf (A cohesion measure for a graph)}{\em
As an illustrating example, suppose $A=(a_{i,j})$ is the $n \times n$ adjacency matrix of a simple undirected graph with $a_{i,j}=1$ if there is an edge between node $i$ and node $j$ and 0 otherwise. Let $k_i=\sum_{j=1}^n a_{i,j}$ be the degree of node $i$ and $m={1 \over 2}\sum_{i=1}^n k_i$ be the total number of edges in the graph.
Then one can simply let $\beta_1(i,j)=2\delta_{i,j}+a_{i,j}$, where $\delta_{i,j}=1$ if $i=j$ and 0 otherwise.
By doing so, we then have the following cohesion measure
\beq{graph1111}
\beta(i,j)=2\delta_{i,j}+a_{i,j}-{{2+k_i} \over n}-{{2+k_j} \over n}+{{2m+2n} \over {n^2}}.
\eeq
We note that such a cohesion measure is known as the deviation to indetermination null model in \cite{campigotto2014generalized}.
}
\eex

\bsec{Conclusions}{conclusion}

In this paper, we developed a mathematical theory for clustering in metric spaces based on distance measures and cohesion measures.
A cluster is defined as a set of data points that are cohesive to themselves.  The hierarchical agglomerative  algorithm
 in Algorithm \ref{alg:hierarchical} was shown to converge with a partition of clusters. Our hierarchical agglomerative  algorithm differs from a standard hierarchical agglomerative  algorithm in two aspects: (i) there is a stopping criterion for our algorithm, and (ii) there is no need to use the greedy selection. We also proposed the $K$-sets algorithm in Algorithm \ref{alg:ksets} based on the concept of triangular distance. Such an algorithm appears to be new. Unlike the Lloyd iteration, it only takes one-step minimization in each iteration and that might give the $K$-sets algorithm the computational advantage over the $K$-medoids algorithms.
 The $K$-sets algorithm was shown to converge with a partition of two clusters when $K=2$. Another interesting finding of the paper is the duality result between a distance measure and a cohesion measure. As such, one can perform clustering either by a distance measure or a cohesion measure. In particular, the dual $K$-sets algorithm in Algorithm \ref{alg:dualksets} converges in the same way as a sequential version of the kernel $K$-means algorithm without the need for the cohesion matrix to {\em positive semi-definite}.

There are several possible extensions for our work:

\noindent (i) {\em Asymmetric distance measure}: One possible extension is to remove the symmetric property in (D3) for a distance measure. Our preliminary result shows that one only needs $d(x,x)=0$ in (D2) and the triangular inequality in (D4) for the $K$-sets algorithm to converge. The key insight for this is that one can replace the original distance measure $d(x,y)$ by a new distance measure $\tilde d(x,y)=d(x,y)+d(y,x)$. By doing so, the new distance measure is symmetric.

\noindent (ii) {\em Distance measure without the triangular inequality}: Another possible extension is to remove the triangular inequality in (D4). However, the $K$-sets algorithm does not work properly in this setting as the triangular distance is no longer nonnegative. In order for the $K$-sets algorithm to converge, our preliminary result shows that one can adjust the value of the triangular distance based on a weaker notion of cohesion measure.
Results along this line will be reported separately.

\noindent (iii) {\em Performance guarantee}: Like the $K$-means algorithm, the output of the $K$-sets algorithm also depends on the initial partition. It would be of interest to see if it is possible to derive performance guarantee for the $K$-sets algorithm (or the optimization problem for the normalized modularity). In particular, the approach by approximation stability in \cite{balcan2013clustering} might be applicable as their threshold graph lemma seems to hold when one replaces the distance from a point $x$ to its center $c$, i.e., $d(x,c)$, by the average distance of a point $x$ to its set, i.e., $\dbar(x,S)$.

\noindent (iv) {\em Local clustering}: The problem of local clustering is to find a cluster that contains a specific point $x$. Since we already define what a cluster is, we may use the hierarchical agglomerative  algorithm
 in Algorithm \ref{alg:hierarchical} to find a cluster that contains $x$. One potential problem of such an approach is the output cluster might be very big. Analogous to the concept of community strength in \cite{chang2013relative}, it would be of interest to define a concept of {\em cluster strength} and stop the agglomerative process when the desired cluster strength can no longer be met.

\noindent (v) {\em Reduction of computational complexity}:
Note that the computation complexity for each iteration within the FOR loop of the $K$-sets algorithm is $O(K n^2)$ as it takes $O(Kn)$ steps to compute the triangular distance for each point and there are $n$ points that need to be assigned in each iteration. To further reduce the computational complexity for such an algorithm, one might exploit the idea of ``sparsity'' and this can be done by the transformation of distance measure.

\section*{Acknowledgement}

This work was supported in part by the Excellent Research Projects of National Taiwan University, under Grant Number AE00-00-04, and in part by National Science Council (NSC), Taiwan, under Grant Numbers NSC102-2221-E-002-014-MY2 and 102-2221-E-007 -006 -MY3.

\bibliographystyle{unsrt}
\bibliography{bibliographyJACM}

\clearpage

\appendix
\section*{Appendix A}

\setcounter{section}{1}

In this section, we prove \rprop{cohesivexy}.

(i)
Since the distance measure $d(\cdot, \cdot)$ is symmetric, we have from \req{coh2222xya} that $\gamma(x,y)=\gamma(y,x)$.
Thus, the cohesion measure between two points is symmetric.

(ii) We note from  \req{coh2222xyb}
that
\bear{coh3333xy}
\gamma(x, x)& =&{1 \over n^2} \sum_{z_2 \in \Omega}\sum_{z_1 \in \Omega}\Big (d(x,z_1)+d(z_2,x) \nonumber \\
&&\quad\quad-d(z_1,z_2)-d(x,x) \Big).
\eear
Since $d(x,x)=0$, we have from the triangular inequality that $\gamma(x,x) \ge 0$.

(iii) Note from \req{coh3333xy} and \req{coh2222xyb} that
\bearn
&&\gamma(x,x)-\gamma(x,y)\\
&&={1 \over n}\sum_{z_2 \in \Omega}\Big (d(z_2,x)-d(x,x) +d(x,y)-d(z_2,y)\Big).
\eearn
Since $d(x,x)=0$, we have from the triangular inequality that
$$\gamma(x,x) \ge \gamma(x,y).$$

(iv) This can be easily verified by summing $y$ in \req{coh2222xya}.

\section*{Appendix B}

In this section, we prove \rthe{clustereq}.

We first list several properties for the average distance that will be used in the proof of \rthe{clustereq}.

\begin{fact}
\label{prop:avg}
\begin{description}
\item[(i)] $\dbar(S_1, S_2) \ge 0$;
\item[(ii)] (Symmetry) $\dbar(S_1, S_2)=\dbar(S_2, S_1)$;
\item[(iii)] (Weighted average) Suppose that $S_2$ and $S_3$ are two disjoint subsets of $\Omega$. Then
\bear{avg2222}
&&\dbar(S_1, S_2 \cup S_3) \nonumber \\
&& ={{|S_2|} \over {|S_2|+{|S_3| }}} \dbar(S_1, S_2)
+{{|S_3|} \over {|S_2|+{|S_3| }}} \dbar(S_1, S_3). \nonumber \\
\eear
\end{description}
\end{fact}

Now we can use the average distance to represent the relative distance and the cohesion measure.

\begin{fact}
\label{prop:repre}
\begin{description}
\item[(i)] From \req{coh2222xya}, one can represent the cohesion measure between two points in terms of the average distance as follows:
\beq{coh1111a}
\gamma(x, y) =\dbar(\Omega, y)+\dbar (x, \Omega)
-\dbar (\Omega, \Omega)-d(x, y).\eeq
\item[(ii)] From \req{coh1111} and  \req{coh1111a},  one can represent the cohesion measure between two sets in terms of the average distance as follows:
\bear{coh1111s}
\gamma(S_1, S_2) &=&|S_1| \cdot |S_2| \cdot \Big (\dbar(\Omega, S_2)+\dbar (S_1, \Omega) \nonumber\\
&&\quad-\dbar (\Omega, \Omega)-\dbar(S_1, S_2) \Big ).
\eear
\item[(iii)] From \req{noravg0000xy}, one can represent the relative distance from $x$ to $y$ in terms of the average distance as follows:
\beq{reldis1111}
\RC(x||y)=d(x,y)-\dbar (x, \Omega).
\eeq
\item[(iv)] From \req{reldis1111} and \req{noravg0000}, one can represent the relative distance from set $S_1$ to another set $S_2$ in terms of the average distance  as follows:
\beq{reldis2222}
\RC(S_1|| S_2)=\dbar(S_1, S_2)-\dbar(S_1, \Omega).
\eeq
\end{description}
\end{fact}

Using the representation of the cohesion measure by the average distance,  we show  some properties for the cohesion measure between two sets in the following proposition.

\bprop{cohfacts}
\begin{description}
\item[(i)] (Symmetry) For any two sets $S_1$ and $S_2$,
\beq{coh1111b}
\gamma(S_1, S_2) =\gamma(S_2, S_1).
\eeq
\item[(ii)] (Zero-sum) Any set $S$ is both cohesive and incohesive to $\Omega$, i.e., $\gamma (\Omega, S)=\gamma(S, \Omega)=0$.
\item[(iii)] (Union of two disjoint sets) Suppose that $S_2$ and $S_3$ are two disjoint subsets of $\Omega$. Then for any set $S_1$
\beq{coh3333}
  \gamma(S_1 , S_2 \cup S_3) = \gamma(S_1 , S_2) +\gamma(S_1 , S_3).
\eeq
\item[(iv)] (Union of two disjoint sets) Suppose that $S_2$ and $S_3$ are two disjoint subsets of $\Omega$. Then for any set $S_1$
\beq{coh3333b}
  \gamma(S_2 \cup S_3,S_1) =\gamma(S_2 , S_1) + \gamma(S_3 , S_1).
\eeq
\item[(v)] Suppose that $S$ is a nonempty set and it is not equal to $\Omega$. Let $S^c=\Omega \backslash S$ be the set of points that are not in $S$.  Then
\beq{coh6666}
\gamma(S,S)=- \gamma (S^c, S)= \gamma(S^c,S^c).
\eeq

\end{description}
\eprop

\bproof
(i) The symmetric property follows from the symmetric property of $\gamma(x,y)$.

(ii) That $\gamma (\Omega, S)=0$ follows trivially from \req{coh1111s}.

(iii) This is a direct consequence of the definition in \req{coh1111}.

(iv) That \req{coh3333b} holds follows from the symmetric property in (i) and the identity in \req{coh3333}.

(v) Note from (ii) of this proposition that
$\gamma (\Omega, S)=\gamma (S\cup S^c,S)=0$.
Using  \req{coh3333b} yields
$$\gamma(S,S)+ \gamma(S^c, S)=0.$$
Thus, we have
\beq{coh6666a}
\gamma(S,S)=-\gamma (S^c, S).
\eeq
From \req{coh6666a}, we also have
\beq{coh6666b}
\gamma(S^c,S^c)=- \gamma (S, S^c).
\eeq
From the symmetric property in (i) of this proposition, we have   $\gamma (S^c, S)=\gamma (S, S^c)$.
As a result of \req{coh6666a} and \req{coh6666b}, we then have
$$\gamma(S,S)= \gamma(S^c,S^c).$$
\eproof



Using the representations of the relative distance and the cohesion measure by the average distance, we show some properties for the relative distance in the following proposition.

\bprop{facts}
\begin{description}
\item[(i)] (Zero-sum) The relative distance from any set $S$ to $\Omega$ is 0, i.e., $\RC(S || \Omega)=0$.
\item[(ii)] (Union of two disjoint sets) Suppose that $S_2$ and $S_3$ are two disjoint subsets of $\Omega$. Then for any set $S_1$
\bear{avg3333}
&&\RC(S_1 || S_2 \cup S_3) ={{|S_2|} \over {|S_2|+{|S_3| }}} \RC(S_1 || S_2) \nonumber\\
&&\quad\quad+{{|S_3|} \over {|S_2|+{|S_3| }}} \RC(S_1 || S_3).
\eear
\item[(iii)] (Union of two disjoint sets) Suppose that $S_2$ and $S_3$ are two disjoint subsets of $\Omega$. Then for any set $S_1$
\bear{avg4444}
&&\RC( S_2 \cup S_3||S_1) ={{|S_2|} \over {|S_2|+{|S_3| }}} \RC(S_2 || S_1) \nonumber\\
&&\quad\quad+{{|S_3|} \over {|S_2|+{|S_3| }}} \RC(S_3 || S_1).
\eear
\item[(iv)] (Reciprocity) For any two sets $S_1$ and $S_2$,
\bear{avg5555}
&&\RC(\Omega || S_2)-\RC(S_1 ||S_2)\nonumber\\
&&=\RC (\Omega || S_1)-\RC(S_2 || S_1).
\eear
\item[(v)] The cohesion measure can be represented by the relative distance as follows:
\beq{avg6666}
\gamma(S_1, S_2)=|S_1| \cdot |S_2| \cdot \Big (\RC(\Omega || S_2)-\RC(S_1 ||S_2)\Big ).
\eeq
\end{description}
\eprop

\bproof
(i) This is trivial from \req{reldis2222}.

(ii) and (iii) follows from \req{avg2222} and \req{reldis2222}.

(iv) Note from the symmetric property of $\dbar(\cdot, \cdot)$ that
\bearn
&&\RC(\Omega || S_2)-\RC(S_1 ||S_2)\\
&&=\dbar(\Omega, S_2)-\dbar (\Omega, \Omega)-(\dbar(S_1, S_2)-\dbar (S_1, \Omega))\\
&&=\dbar(\Omega, S_1)-\dbar (\Omega, \Omega)-(\dbar(S_2, S_1)-\dbar (S_2, \Omega))\\
&&=\RC (\Omega || S_1)-\RC(S_2 || S_1)
\eearn

(v) This is a direct consequence of \req{coh1111s} and \req{reldis2222}.
\eproof

%
\bproof {\bf (\rthe{clustereq})}
(i) $\Rightarrow$ (ii): If $\gamma(S,S) \ge 0$, we then have
 from \req{coh6666} that
\beq{coh6666p}
\gamma(S^c,S^c)=\gamma(S,S)\ge 0 .
\eeq

(ii) $\Rightarrow$ (iii): If $\gamma(S^c,S^c) \ge 0$, then
it follows from \req{coh6666} that
\beq{coh6666q}
\gamma (S^c, S)=-\gamma(S^c,S^c) \le 0.
\eeq

(iii) $\Rightarrow$ (iv): If $\gamma(S^c,S) \le 0$, then we have from \req{coh6666} that
$\gamma(S,S)=- \gamma(S^c,S)  \ge 0$. Thus, $\gamma(S,S) \ge \gamma(S,S^c)$.

(iv) $\Rightarrow$ (v): If $\gamma(S,S) \ge \gamma(S,S^c)$, then
it follows from \req{coh6666q} that
$$\gamma(S,S) \ge \gamma(S,S^c)=-\gamma(S,S).$$
This then leads to $\gamma(S,S) \ge 0$.
From \req{coh1111s}, we know that
\beq{cluster2222}
\gamma(S, S)=|S|^2 \cdot \Big (2\dbar (S, \Omega)-\dbar(\Omega, \Omega)-\dbar (S,S) \Big )\ge 0.
\eeq
Thus, $2\dbar (S, \Omega)-\dbar(\Omega, \Omega)-\dbar (S,S)\ge 0$.

(v) $\Rightarrow$ (vi): Note from \req{reldis2222} that
\bearn
&&\RC(\Omega||S) - \RC(S||S)\nonumber \\
&&=\dbar(\Omega, S)-\dbar(\Omega, \Omega)-(\dbar(S,S)-\dbar(S, \Omega))\\
&&=2\dbar (S, \Omega)-\dbar(\Omega, \Omega)-\dbar (S,S).
\eearn
Thus, if $2\dbar (S, \Omega)-\dbar(\Omega, \Omega)-\dbar (S,S)\ge 0$, then $\RC(\Omega||S) \ge \RC(S||S)$.

(vi) $\Rightarrow$ (vii): From \req{avg4444} in \rprop{facts}(iii), we have
$$\RC(\Omega|| S)=\RC(S \cup S^c||S) ={{|S|} \over n} \RC(S ||S) + {{|S^c|} \over n} \RC(S^c ||S).$$
Thus,
$$\RC(\Omega|| S)- \RC(S||S)={{|S^c|} \over n} \Big (\RC(S^c||S) -\RC(S||S) \Big ).$$
Clearly, if $\RC(\Omega||S) \ge \RC(S||S)$, then $\RC(S^c||S) \ge \RC(S||S)$.

(vii) $\Rightarrow$ (viii):
 Note from \req{avg2222} and \req{reldis2222} that
\bearn
&&\RC(S^c||S)-\RC(S||S) \\
&&=\dbar(S^c, S)-\dbar(S^c, \Omega)-(\dbar(S,S)-\dbar(S,\Omega))\\
&&=\dbar(S^c, S)-\dbar(S^c, S \cup S^c)-\dbar(S,S)+\dbar(S,S \cup S^c)  \\
&&={{(n-|S|)} \over {n}}\Big (2\dbar(S, S^c)- \dbar(S,S)-\dbar(S^c,S^c)\Big ).
\eearn
Thus, if $\RC(S^c||S) \ge \RC(S||S)$, then $2\dbar(S, S^c)- \dbar(S,S)-\dbar(S^c,S^c) \ge 0$.

(viii) $\Rightarrow$ (ix): Note from \req{avg2222} and \req{reldis2222}  that
\bearn
&&\RC(S||S^c) - \RC(\Omega|| S^c) \\
&&=\dbar (S, S^c)-\dbar(S, \Omega)-(\dbar (\Omega, S^c)-\dbar(\Omega, \Omega))\\
&&=\dbar (S, S^c)-\dbar (S,S \cup S^c) - \dbar(S \cup S^c, S^c)\\
&&\quad\quad+\dbar(S \cup S^c, S \cup S^c)   \\
&&={{|S| \times (n-|S|)} \over {n^2}}\Big (2\dbar(S, S^c)- \dbar(S,S)-\dbar(S^c,S^c)\Big ).
\eearn
Thus, if $2\dbar(S, S^c)- \dbar(S,S)-\dbar(S^c,S^c) \ge 0$, then $\RC(S||S^c) \ge \RC(\Omega|| S^c)$.

(ix) $\Rightarrow$ (x): Note from \req{avg5555} in \rprop{facts}(iv) that
$$\RC(S^c||S) - \RC(\Omega||S)= \RC(S||S^c) - \RC(\Omega|| S^c).$$
Thus, if $\RC(S||S^c) \ge \RC(\Omega|| S^c)$, then $\RC(S^c||S) \ge \RC(\Omega||S)$.

(x) $\Rightarrow$ (i): Note from \req{coh6666} and \req{avg6666} that
$$\gamma(S,S) =-\gamma(S, S^c)=-|S| \cdot |S^c| \cdot \Big ( \RC(\Omega||S)-\RC(S^c||S)\Big ).$$
Thus, if $\RC(S^c||S) \ge \RC(\Omega||S)$, then $\gamma(S,S) \ge 0$.

\eproof

\section*{Appendix C}

In this section, we prove \rthe{community}.

(i) We prove this by induction. Since $\gamma(x,x)\ge 0$, every point is a cluster by itself. Thus, all the initial $n$ sets are disjoint clusters.
Assume that all the remaining sets are clusters as our induction hypothesis. In each iteration, we merge two disjoint cohesive clusters.
Suppose that $S_i$ and $S_j$ are merged to form $S_k$. It then follows from \req{coh3333} and \req{coh3333b} in \rprop{cohfacts}(iii) and (iv) that
\beq{random4400b}
\gamma(S_k, S_k)= \gamma(S_i, S_i)+2 \gamma (S_i, S_j) + \gamma (S_j, S_j).
\eeq
As both $S_i$ and $S_j$ are clusters from our induction hypothesis, we have $\gamma(S_i, S_i) \ge 0$ and $\gamma (S_j, S_j)\ge 0$.
Also, since $S_i$ and $S_j$ are cohesive, i.e., $\gamma (S_i, S_j)\ge 0$,  we then have from \req{random4400b} that
$\gamma(S_k, S_k) \ge 0$ and the set $S_k$ is also a cluster.

(ii) To  see that the modularity index is non-decreasing in every iteration, note from \req{random4400b} and $\gamma (S_i, S_j)\ge 0$ that
$$\gamma(S_k, S_k)\ge  \gamma(S_i, S_i) +\gamma (S_j, S_j).$$
As such, the algorithm converges to a local optimum.

\section*{Appendix D}

In this appendix, we prove \rlem{totalstr}.

(i) From the triangular inequality, it is easy to see from the definition of the triangular distance in \req{triang1111}
that $\Delta(x,S) \ge 0$.
Note that
$${1 \over {|S|^2}}\sum_{z_1 \in S} \sum_{z_2 \in S}d(x,z_1)
={1 \over {|S|}}\sum_{z_1 \in S}d(x,z_1)= \dbar(\{x\}, S).$$
Similarly,
$${1 \over {|S|^2}}\sum_{z_1 \in S} \sum_{z_2 \in S}d(x,z_2)
= \dbar(\{x\}, S).$$
Thus, the triangular distance in \req{triang1111} can also be written as $\Delta(x, S)=2\dbar(\{x\}, S) - \dbar(S,S)$.

(ii)
Recall from \req{coh1111s} that
\bearn
\gamma(S_1, S_2)
&=&|S_1| \cdot |S_2| \cdot \Big (\dbar(\Omega, S_2)+\dbar (S_1, \Omega)\\
&&\quad\quad
-\dbar (\Omega, \Omega)-\dbar(S_1, S_2) \Big ).
\eearn
Thus,
\bearn
&&\gamma(x,x)-{2 \over |S|} \gamma(\{x\},S)+ {1 \over {|S|^2}}\gamma(S,S) \\
&&=2\dbar(\{x\}, \Omega)-\dbar(\Omega, \Omega)-2\Big (\dbar(\{x\}, \Omega)+\dbar (S, \Omega)\nonumber\\
&&\quad\quad
-\dbar (\Omega, \Omega)-\dbar(\{x\}, S) \Big ) \\
&&\quad\quad+\Big (2\dbar(\Omega, S)
-\dbar (\Omega, \Omega)-\dbar(S, S) \Big )\\
&&=2\dbar(\{x\}, S) - \dbar(S,S)=\Delta(x, S),
\eearn
where we use (i) of the lemma in the last equality.

(iii) Note from (ii) that
\bear{totalstr3333b}
&&\sum_{k=1}^K \sum_{x \in S_k} \Delta(x, S_k) \nonumber \\
&&=\sum_{k=1}^K \sum_{x \in S_k} \Big ( \gamma(x,x)-{2 \over |S_k|} \gamma(\{x\},S_k)\nonumber\\
&&\quad\quad\quad\quad+ {1 \over {|S_k|^2}}\gamma(S_k,S_k)\Big) \nonumber \\
&&=\sum_{k=1}^K \sum_{x \in S_k}\gamma(x,x)-\sum_{k=1}^K  {1 \over {|S_k|}}\gamma(S_k,S_k) \nonumber\\
&&=\sum_{x \in \Omega}\gamma(x,x)-R .
\eear

(iv) From (i) of this lemma, we have
$$\sum_{k=1}^K \sum_{x \in S_k} \Delta(x, S_k) = \sum_{k=1}^K \sum_{x \in S_k} (2\dbar(\{x\}, S_k) - \dbar(S_k,S_k)).
$$
Observe that
$$\sum_{x \in S_k} \dbar(\{x\}, S_k)=|S_k| \dbar (S_k, S_k).$$
Thus,
\bearn
&&\sum_{k=1}^K \sum_{x \in S_k} \Delta(x, S_k) = \sum_{k=1}^K |S_k| \cdot (2\dbar (S_k, S_k)-\dbar(S_k,S_k))\\
&&=\sum_{k=1}^K |S_k| \cdot \dbar(S_k,S_k)=\sum_{k=1}^K \sum_{x \in S_k}\dbar(\{x\}, S_k)\\
&&=\sum_{x \in \Omega}\dbar (x, S_{c(x)}).
\eearn

\section*{Appendix E}

In this section, we prove \rthe{triangular}.
For this, we need to prove the following two inequalities.

\blem{twoinequalities}
For any set $S$ and any point $x$ that is not in $S$,
\beq{totalstr4444}
\sum_{y \in S\cup \{x\}}\Delta(y, S\cup \{x\})\le \sum_{y \in S\cup \{x\}}\Delta(y, S),
\eeq
and
\beq{totalstr5555}
\sum_{y \in S}\Delta(y, S) \le \sum_{y \in S}\Delta(y, S\cup \{x\}).
\eeq
\elem

\bproof
We first show that for any set $S$ and any point $x$ that is not in $S$,
\beq{avg2255}
\dbar(S\cup \{x\}, S\cup \{x\})-2\dbar(S\cup \{x\},S)+ \dbar(S,S) \le 0.
\eeq
From the symmetric property in Fact \ref{prop:avg}(ii) and the weighted average property in Fact \ref{prop:avg}(iii), we have
\bearn
&&\dbar(S\cup \{x\}, S\cup \{x\})={{|S|^2}  \over {(|S|+1)^2}}\dbar (S,S)\\
&&\quad+{{2|S|}  \over {(|S|+1)^2}}\dbar (\{x\},S)
+{{1}  \over {(|S|+1)^2}}\dbar (\{x\},\{x\}),
\eearn
and
$$\dbar(S\cup \{x\},S)={{|S|} \over {|S|+1}} \dbar (S,S)+ {{1} \over {|S|+1}} \dbar (\{x\},S).$$
Note that
$$\dbar (\{x\},\{x\})=d(x,x)=0.$$
Thus,
\bearn
&&\dbar(S\cup \{x\}, S\cup \{x\})-2\dbar(S\cup \{x\},S)+ \dbar(S,S) \\
&&={{1}  \over {(|S|+1)^2}} \Big (\dbar(S,S)-2\dbar(\{x\}, S)\Big ) \le 0,
\eearn
where we use \req{triang1111b} in the last inequality.

Note from \req{triang1111b} that
\bear{totalstr5555b}
&&\sum_{y \in S_2}\Delta(y, S_1) =\sum_{y \in S_2}(2\dbar(\{y\}, S_1) - \dbar(S_1,S_1))\nonumber\\
&&= |S_2| \cdot \Big (2 \dbar(S_1, S_2)-\dbar(S_1,S_1)\Big ) .
\eear
Using \req{totalstr5555b} yields
\bear{totalstr6666}
&&\sum_{y \in S\cup \{x\}}\Delta(y, S\cup \{x\}) -\sum_{y \in S\cup \{x\}}\Delta(y, S) \nonumber \\
&&=|S\cup \{x\}| \cdot \Big (2 \dbar(S\cup \{x\}, S\cup \{x\})\nonumber\\
&&\quad\quad-\dbar(S\cup \{x\},S\cup \{x\})\Big ) \nonumber \\
&&\quad\quad - |S\cup \{x\}| \cdot \Big (2 \dbar(S, S\cup \{x\})-\dbar(S,S)\Big ) \nonumber \\
&&=|S\cup \{x\}| \cdot \Big (\dbar(S\cup \{x\}, S\cup \{x\})\nonumber\\
&&\quad\quad-2\dbar(S, S\cup \{x\})+ \dbar(S,S) \Big ).
\eear
As a result of \req{avg2255}, we then have
$$\sum_{y \in S\cup \{x\}}\Delta(y, S\cup \{x\}) -\sum_{y \in S\cup \{x\}}\Delta(y, S)
\le 0.$$

Similarly, using \req{totalstr5555b} and \req{avg2255} yields
\bear{totalstr7777}
&&\sum_{y \in S}\Delta(y, S)-\sum_{y \in S}\Delta(y, S\cup \{x\}) \nonumber \\
&&=|S| \cdot \Big (2 \dbar(S, S)-\dbar(S,S)\Big )  \nonumber \\
&&\quad\quad -|S| \cdot \Big (2 \dbar(S\cup \{x\}, S)-\dbar(S\cup \{x\},S\cup \{x\})\Big ) \nonumber\\
&&=|S| \cdot \Big (\dbar(S\cup \{x\}, S\cup \{x\})\nonumber\\
&&\quad\quad-2\dbar(S\cup \{x\},S)+ \dbar(S,S) \Big )\nonumber\\
&&
\le 0
\eear
\eproof

\bproof {\bf (\rthe{triangular})}
(i)
Let $S_k$ (resp. $S^\prime_k$), $k=1,2, \ldots, K$, be the partition before (resp. after) the change.
Also let $c(x)$ be the index of the set to which $x$ belongs.
Suppose that  $\Delta(x_i, S_{k^*}) < \Delta(x_i, S_{c(x_i)})$ and $x_i$ is moved from $S_{c(x_i)}$ to $S_{k^*}$ for some point $x_i$ and some $k^*$.
In this case, we have $S^\prime_{k^*}=S_{k^*} \cup\{x_i\}$, $S^\prime_{c(x_i)}=S^\prime_{c(x_i)}\backslash \{x\}$ and $S^\prime_k=S_k$ for all $k \ne c(x_i), k^*$.
It then follows from $\Delta(x_i, S_{k^*}) < \Delta(x_i, S_{c(x_i)})$ that
\bear{triang4444}
&&\sum_{k=1}^K \sum_{x \in S_k}\Delta(x, S_k) \nonumber \\
&&=\sum_{k \ne c(x_i), k^*} \sum_{x \in S_k}\Delta(x, S_k)\nonumber \\
&&\quad+ \sum_{x \in S_{c(x_i)} }\Delta(x, S_{c(x_i)})+
\sum_{x \in S_{k^*}} \Delta(x, S_{k^*}) \nonumber\\
&&=\sum_{k \ne c(x_i), k^*} \sum_{x \in S_k}\Delta(x, S_k)+ \sum_{x \in S_{c(x_i)}\backslash \{x_i\} }\Delta(x, S_{c(x_i)})\nonumber \\
&&\quad+\Delta(x_i, S_{c(x_i)})+
\sum_{x \in S_{k^*}} \Delta(x, S_{k^*}) \nonumber\\
&&> \sum_{k \ne c(x_i), k^*} \sum_{x \in S_k}\Delta(x, S_k)+ \sum_{x \in S_{c(x_i)}\backslash \{x_i\} }\Delta(x, S_{c(x_i)})\nonumber \\
&&\quad+\Delta(x_i, S_{k^*})+
\sum_{x \in S_{k^*}} \Delta(x, S_{k^*}) \nonumber\\
&&=\sum_{k \ne c(x_i), k^*} \sum_{x \in S_k^\prime}\Delta(x, S_k^\prime)+ \sum_{x \in S_{c(x_i)}\backslash \{x_i\} }\Delta(x, S_{c(x_i)})\nonumber \\
&&\quad+
\sum_{x \in S_{k^*}\cup\{x_i\}} \Delta(x, S_{k^*}),
\eear
where we use the fact that $S^\prime_k=S_k$ for all $k \ne c(x_i), k^*$, in the last equality.
From \req{totalstr4444} and $S^\prime_{k^*}=S_{k^*} \cup\{x_i\}$, we know that
\bear{triang5555}
\sum_{x \in S_{k^*}\cup\{x_i\}} \Delta(x, S_{k^*}) &\ge&
\sum_{x \in S_{k^*}\cup\{x_i\}} \Delta(x, S_{k^*}\cup\{x_i\})\nonumber\\
&=&
\sum_{x \in S^\prime_{k^*}} \Delta(x, S^\prime_{k^*}).
\eear
Also, it follows from \req{totalstr5555} and $S^\prime_{c(x_i)}=S^\prime_{c(x_i)}\backslash \{x_i\}$ that
\bear{triang6666}
\sum_{x \in S_{c(x_i)}\backslash \{x_i\} }\Delta(x, S_{c(x_i)}) &\ge&
\sum_{x \in S_{c(x_i)}\backslash \{x_i\} }\Delta(x, S_{c(x_i)}\backslash \{x_i\} )
\nonumber\\&=&\sum_{x \in S^\prime _{c(x_i)} }\Delta(x, S^\prime_{c(x_i)}).
\eear
Using \req{triang5555} and \req{triang6666} in \req{triang4444} yields
\bear{triang7777}
&&\sum_{k=1}^K \sum_{x \in S_k}\Delta(x, S_k) \nonumber \\
&&> \sum_{k \ne c(x_i), k^*} \sum_{x \in S_k^\prime}\Delta(x, S_k^\prime)+\sum_{x \in S^\prime_{k^*}} \Delta(x, S^\prime_{k^*})\nonumber\\
&&\quad\quad+\sum_{x \in S^\prime _{c(x_i)} }\Delta(x, S^\prime_{c(x_i)})\nonumber \\
&&=\sum_{k=1}^K \sum_{x \in S^\prime_k}\Delta(x, S^\prime_k).
\eear

In view of \req{totalstr3333} and \req{triang7777}, we then conclude that the normalized modularity is increasing
when there is a change. Since there is only a finite number of partitions for $\Omega$, the algorithm thus converges to a local optimum of the normalized modularity.

(ii)
The algorithm converges when there are no further changes. As such, we know for any $x \in S_i$ and $j \ne i$,
$\Delta(x, S_i) \le \Delta (x, S_j)$. Summing up all the points $x \in S_i$ and using \req{triang1111b} yields
\bear{twoc2255}
&&0 \ge \sum_{x \in S_i}\Big (\Delta(x, S_i)- \Delta (x, S_j) \Big )\nonumber \\
&&=\sum_{x \in S_i}\Big (2\dbar (\{x\}, S_i)-\dbar(S_i, S_i)\Big) \nonumber\\
&&\quad\quad-\sum_{x \in S_i}\Big (2\dbar (\{x\}, S_j)-\dbar(S_j, S_j)\Big) \nonumber\\
&&=|S_i|\cdot \Big (\dbar(S_i,S_i)-2\dbar(S_i, S_j)+\dbar(S_j, S_j)\Big ).
\eear
When the two sets $S_i$ and $S_j$ are viewed in isolation (by removing the data points not in $S_i \cup S_j$ from $\Omega$), we have $S_j=S_i^c$.
Thus,
$$\dbar(S_i,S_i)-2\dbar(S_i, S_i^c)+\dbar(S_i^c, S_i^c) \le 0.$$
As a result of \rthe{clustereq}(viii), we conclude that $S_i$ is a cluster when the two sets $S_i$ and $S_j$ are viewed in isolation.
Also,  \rthe{clustereq}(ii) implies that $S_j=S_i^c$ is also a cluster when the two sets $S_i$ and $S_j$ are viewed in isolation.
\eproof

\section*{Appendix F}

In this section, we prove \rlem{csimdist}.

We first show (C1).
Since $d(x,y)=d(y,x)$ for all $x \ne y$, we have from \req{csim7777} that
$\beta(x,y)=\beta(y,x)$ for all $x \ne y$.

To verify (C2), note from \req{csim7777} that
\bear{csim3333b}
\sum_{y \in \Omega}\beta(x,y)&=&
{1 \over n}\sum_{y \in \Omega}\sum_{z_2 \in \Omega} d(z_2,y)+\sum_{z_1 \in \Omega}d(x,z_1) \nonumber\\
&-&{1 \over n} \sum_{z_2 \in \Omega}\sum_{z_1 \in \Omega}d(z_2,z_1)-\sum_{y \in \Omega}d(x,y)\nonumber\\
&=&0.
\eear

Now we show (C3). Note from \req{csim7777} that
\bear{csim4444b}
&&
\beta(x,x)+\beta (y,z) -\beta (x,z)-\beta (x, y)\nonumber\\
&&=-d(x,x)-d(y,z) +d(x,z)+d(x, y).
\eear
Since $d(x,x)=0$, it then follows from the triangular inequality for $d(\cdot,\cdot)$ that
$$\beta(x,x)+\beta (y,z) -\beta (x,z)-\beta (x, y) \ge 0.$$

\section*{Appendix G}

In this section, we prove \rlem{cinduced}.

Clearly, $d(x,x)=0$ from \req{cind1111} and thus (D2) holds trivially.
That  (D3) holds follows from the symmetric property in (C1) of \rdef{cmeasures}.

To see (D1), choosing $z=y$ in \req{cmeas1111} yields
$$0 \le \beta(x,x)+\beta (y,y) -\beta (x,y)-\beta (x, y) =2 d(x,y).$$

For the triangular inequality in (D4), note from \req{cind1111} and \req{cmeas1111} in (C3) that
\bearn
&&  d(x,z)+ d(z,y)-d(x,y) \\
&&={{(\beta(x,x)+\beta(z,z))}\over 2} -\beta(x,z)+{{(\beta(z,z)+\beta(y,y))} \over 2}\\
&&\quad\quad\quad -\beta(z,y)-{{(\beta(x,x)+\beta(y,y))} \over 2} +\beta(x,y)\\
&&=\beta(z,z)+\beta(x,y)-\beta (z,x)-\beta(z,y) \ge 0.
\eearn

\section*{Appendix H}

In this section, we prove \rthe{duality}.

We first show that $d^{**}(x,y)=d(x,y)$ for a distance measure $d(\cdot,\cdot)$.
Note from \req{dual1111} and $d(x,x)=0$ that
\bear{dual1111x}
&&d^*(x,x)={1 \over n}\sum_{z_2 \in \Omega} d(z_2,x)+{1 \over n} \sum_{z_1 \in \Omega}d(x,z_1)\nonumber\\
&&\quad-{1 \over n^2} \sum_{z_2 \in \Omega}\sum_{z_1 \in \Omega}d(z_2,z_1).
\eear
From the symmetric property of $d(\cdot,\cdot)$, it then follows that
\beq{dual1111xx}
d^*(x,x)={2 \over n} \sum_{z_1 \in \Omega}d(x,z_1)-{1 \over n^2} \sum_{z_2 \in \Omega}\sum_{z_1 \in \Omega}d(z_2,z_1).
\eeq
Similarly,
\beq{dual1111yy}
d^*(y,y)={2 \over n}\sum_{z_2 \in \Omega} d(z_2,y)-{1 \over n^2} \sum_{z_2 \in \Omega}\sum_{z_1 \in \Omega}d(z_2,z_1).
\eeq
Using \req{dual1111}, \req{dual1111xx} and \req{dual1111yy} in \req{dual2222} yields
\beq{dual2255}
d^{**}(x,y)=(d^*(x,x)+d^*(y,y))/2 -d^*(x,y) =d(x,y).
\eeq

Now we show that $\beta^{**}(x,y)=\beta(x,y)$ for a cohesion measure $\beta(\cdot,\cdot)$.
Note from \req{dual2222} that
\bear{dual3333}
&&\beta^*(z_2,y)+\beta^*(x,z_1)-\beta^*(z_1,z_2)-\beta^*(x,y) \nonumber\\
&&=-\beta(z_2,y)-\beta(x,z_1)+\beta(z_1,z_2)+\beta(x,y).
\eear
Also, we have from \req{dual1111} that
\bear{dual4444}
&&\beta^{**}(x,y)\nonumber\\
&&={1 \over n}\sum_{z_2 \in \Omega} \beta^*(z_2,y)+{1 \over n} \sum_{z_1 \in \Omega}\beta^*(x,z_1)\nonumber\\
&&\quad\quad-{1 \over n^2} \sum_{z_2 \in \Omega}\sum_{z_1 \in \Omega}\beta^*(z_2,z_1)-\beta^*(x,y)\nonumber\\
&&={1 \over {n^2}}\sum_{z_2 \in \Omega} \sum_{z_1 \in \Omega} \Big (\beta^*(z_2,y)+\beta^*(x,z_1)\nonumber\\
&&\quad\quad-\beta^*(z_1,z_2)-\beta^*(x,y)\Big ).
\eear
Using \req{dual3333} in \req{dual4444} yields
\bear{dual5555}
&&\beta^{**}(x,y)\nonumber\\
&&={1 \over {n^2}}\sum_{z_2 \in \Omega} \sum_{z_1 \in \Omega}\Big (\beta(x,y)+\beta(z_1,z_2)\nonumber\\
&&\quad\quad-\beta(x,z_1)-\beta(z_2,y)\Big)\nonumber\\
&&=\beta(x,y)+{1 \over {n^2}}\sum_{z_2 \in \Omega} \sum_{z_1 \in \Omega}\beta(z_1,z_2)-{1 \over n}\sum_{z_1 \in \Omega}\beta(x,z_1)
\nonumber\\
&&\quad\quad-{1 \over n}\sum_{z_1 \in \Omega}\beta(z_2,y).
\eear
Since $\beta(\cdot,\cdot)$ is a cohesion measure that satisfies (C1)--(C3), we have from (C1) and (C2) that the last three terms in \req{dual5555} are all equal to 0. Thus,
$\beta^{**}(x,y)=\beta(x,y)$.

\section*{Appendix I}

In this section, we prove \rprop{csim}.

We first show (C1).
Since $\beta_1(x,y)=\beta_0(x,y)$ for all $x \ne y$, we have from the symmetric property of $\beta_0(\cdot,\cdot)$ that $\beta_1(x,y)=\beta_1(y,x)$ for all $x \ne y$.
In view of \req{csim2222}, we then also have $\beta(x,y)=\beta(y,x)$ for all $x \ne y$.

To verify (C2), note from \req{csim2222} that
\bear{csim3333}
\sum_{y \in \Omega}\beta(x,y)&=&
\sum_{y \in \Omega} \beta_1(x,y)- {1 \over n}\sum_{y \in \Omega}\sum_{z_1 \in \Omega}\beta_1(z_1,y) \nonumber
\\
&-&\sum_{z_2 \in \Omega}\beta_1(x,z_2) +{1 \over {n}}\sum_{z_1 \in \Omega}\sum_{z_2 \in \Omega}\beta_1(z_1,z_2)\nonumber \\
&=&0.
\eear

Now we show (C3). Note from \req{csim2222} that
\bear{csim4444}
&&
\beta(x,x)+\beta (y,z) -\beta (x,z)-\beta (x, y)\nonumber\\
&&=\beta_1(x,x)+\beta_1 (y,z) -\beta_1 (x,z)-\beta_1 (x, y).
\eear
It then follows from \req{csim1111} that for all $x \ne y \ne z$
that
\beq{csim4455}
\beta(x,x)+\beta (y,z) -\beta (x,z)-\beta (x, y) \ge 0.
\eeq
If either $x=y$ or $x=z$, we also have
$$\beta(x,x)+\beta (y,z) -\beta (x,z)-\beta (x, y) = 0.$$
Thus, it remains to show the case that $y=z$ and $x \ne y$. For this case, we need to show that
\beq{csim5555}\beta(x,x)+\beta (y,y) -\beta (x,y)-\beta (x, y) \ge 0.
\eeq
Note from \req{csim4455} that
\bear{csim4466}
\beta(y,y)+\beta (x,z) -\beta (y,z)-\beta (y, x) \ge 0.
\eear
Summing the two inequalities in \req{csim4455} and \req{csim4466} and using the symmetric property of $\beta(\cdot,\cdot)$ yields
the desired inequality in \req{csim5555}.

\end{document}